%% file: main.tex
\documentclass{article}

\input{setup}
\usepackage{fontawesome5}
\begin{document}

%% ============================================================
%% Title and Authors
%% ============================================================

\title{\textsc{AsiEvolve}: Scientific Discovery with Deep Cognition}

\title{\textsc{Asi-Evolve}: Agentic Discovery with Evolved Cognition}

\title{\textsc{Asi-Evolve}: Agentic Discovery in Long-Horizon Real-World Research}

\title{\textsc{Asi-Evolve}: Automating Long-Horizon Real-World Scientific Research}

\title{\textsc{daVinci-Evolve}: Automating Long-Horizon Real-World Scientific Research}

\title{ASI-Evolve: Long-Horizon AI Research via Agentic Evolution}

\title{ASI-Evolve: AI Accelerates AI}

% Author list - retain lab affiliations
\author[1,2,3]{Weixian Xu\textsuperscript{‡}}
\author[1,2,3]{Tiantian Mi\textsuperscript{*}}
\author[1,2,3]{Yixiu Liu\textsuperscript{*}}
\author[2,3]{Yang Nan\textsuperscript{*}}
\author[2]{Zhimeng Zhou\textsuperscript{*}}
\author[1,2,3]{Lyumanshan Ye}
\author[1,2,3]{Lin Zhang}
\author[2]{Yu Qiao}
\author[1,2,3]{Pengfei Liu\textsuperscript{†}}
\affil{SJTU \quad \textsuperscript{2}SII \quad \textsuperscript{3}GAIR}

\maketitle

%% ============================================================
%% Header Configuration for Title Page
%% ============================================================
\pagestyle{fancy}
\thispagestyle{fancy}
\fancyhead{}
\lhead{
  \raisebox{-0.3cm}{\includegraphics[height=0.95cm]{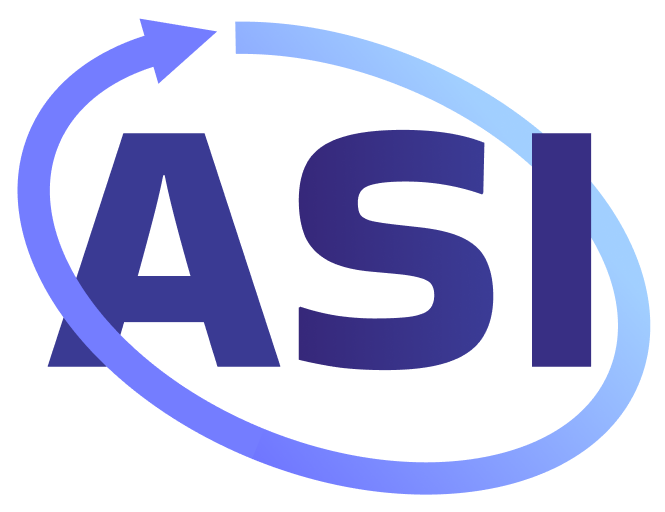}}
}
\rhead{%
  \raisebox{-0.2cm}{\includegraphics[height=0.7cm]{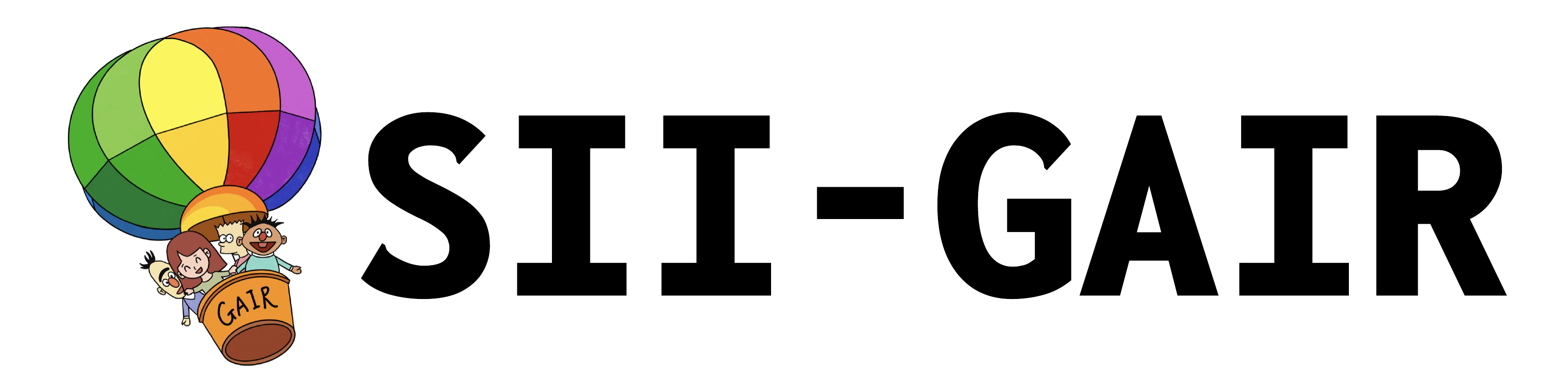}}%
}
\renewcommand{\headrulewidth}{0pt}
\setlength{\headsep}{2mm}

%% ============================================================
%% Author Footnotes
%% ============================================================
\renewcommand{\thefootnote}{}
\footnotetext{‡ Leading author.}
\footnotetext{* Core contributors.}
\footnotetext{† Corresponding author.}
\vspace{-28pt}
\vspace{15pt}
\begin{abstract}

Can AI accelerate the development of AI itself? While recent agentic systems have shown strong performance on well-scoped tasks with rapid feedback, it remains unclear whether they can tackle the costly, long-horizon, and weakly supervised research loops that drive real AI progress. We present \modelname, an agentic framework for AI-for-AI research that closes this loop through a learn–design–experiment–analyze cycle. \modelname augments standard evolutionary agents with two key components: a cognition base that injects accumulated human priors into each round of exploration, and a dedicated analyzer that distills complex experimental outcomes into reusable insights for future iterations.
To our knowledge, \modelname is the first unified framework to demonstrate AI-driven discovery across three central components of AI development: \textbf{data, architectures, and learning algorithms}. In neural architecture design, it discovered \textbf{105 SOTA linear attention architectures}, with the best discovered model surpassing DeltaNet by +0.97 points, nearly 3× the gain of recent human-designed improvements. In pretraining data curation, the evolved pipeline improves average benchmark performance by +3.96 points, with gains \textbf{exceeding 18 points on MMLU}. In reinforcement learning algorithm design, discovered algorithms outperform GRPO by up to \textbf{+12.5 points on AMC32, +11.67 points on AIME24, and +5.04 points on OlympiadBench}. We further provide initial evidence that this AI-for-AI paradigm can transfer beyond the AI stack through experiments in \textbf{mathematics} and \text{biomedicine}.
Together, these results suggest that \modelname represents a promising step toward enabling AI to accelerate AI across the foundational stages of development, offering early evidence for the feasibility of closed-loop AI research. The \modelname is fully open-sourced at \url{https://github.com/GAIR-NLP/ASI-Evolve}.
% \mtt{Perhaps we need some conservative wording.}

% Existing agentic systems have demonstrated strong capabilities on well-scoped scientific tasks with rapid feedback cycles, yet whether AI can tackle large-scale, long-horizon AI research tasks to accelerate its own field remains largely unexplored. To this end, we present \modelname, which introduces a dedicated analyzer module and human cognition priors into the agentic loop: the system learns from accumulated experience and domain priors, designs new solutions, runs experiments to obtain results, and distills insights through analysis to produce new experience, forming a continuously self-improving cycle. We apply \modelname to AI-for-AI research across the three core pillars of AI development---model architecture, data preparation, and training algorithm: in neural architecture design, our best model outperforms DeltaNet with nearly 3$\times$ the gain of recent human-designed SOTA; in pretraining data curation, evolved strategies boost average benchmark scores by 3.96 points with gains exceeding 18 points on MMLU; in reinforcement learning algorithm design, discovered algorithms consistently outperform GRPO across mathematics and coding tasks. We further validate the framework through comparison and ablation studies, and demonstrate that solutions discovered through AI-for-AI research are real-world applicable via experiments in the biomedical domain. These results demonstrate \modelname's capability of using AI to accelerate its own research progress and tackle long-horizon scientific tasks.

\vspace{1.25em}
\begin{figure}[htbp]
    \centering
\includegraphics[width=0.8\textwidth]{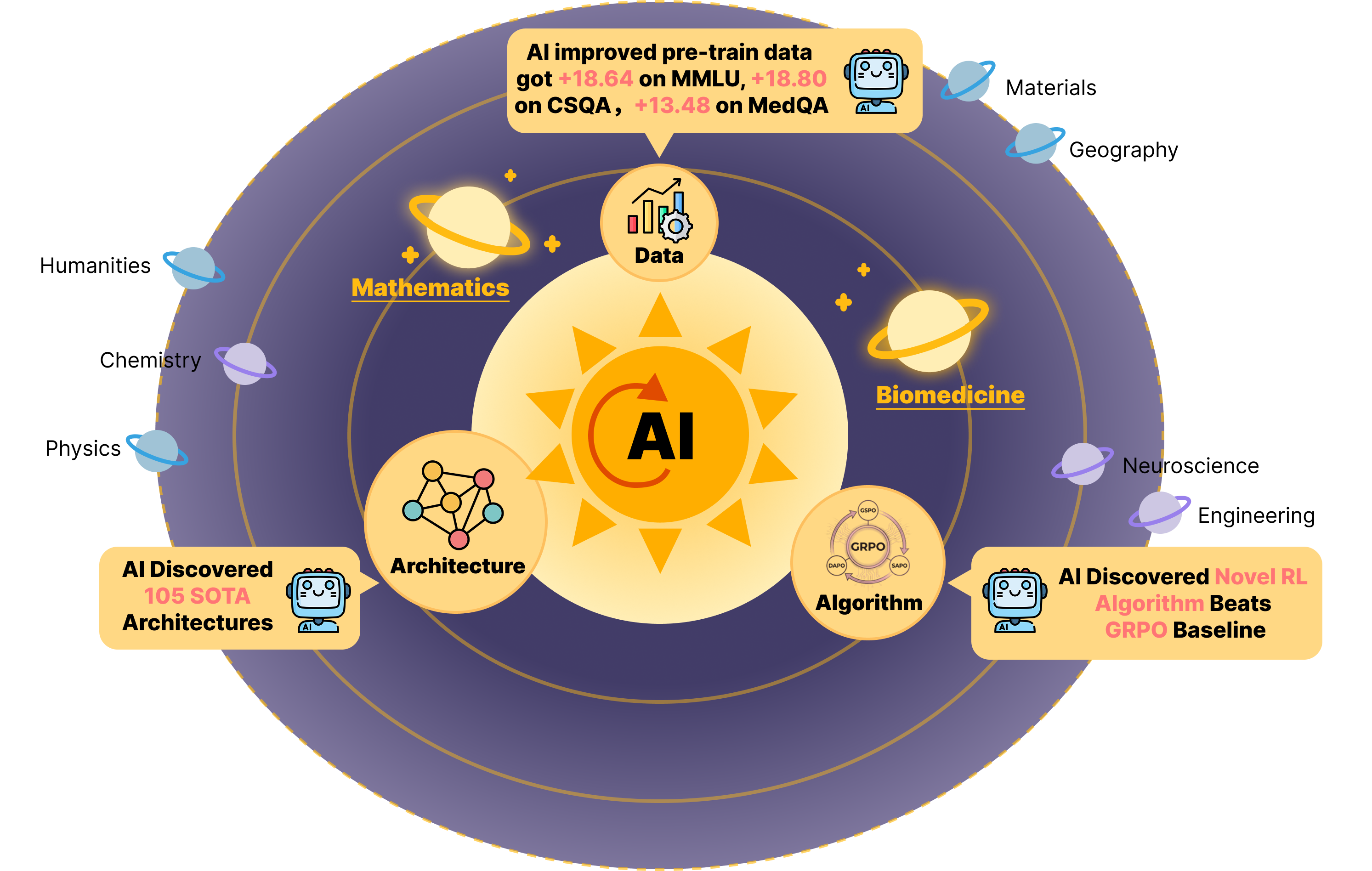}
\end{figure}

\end{abstract}

%% ============================================================
%% Teaser Figure
%% ============================================================

% \begin{figure}[h]
%     \centering
%     \includegraphics[width=0.9\linewidth]{figures/teaser.png}
%     \caption{ASI-Evolve: an evolution framework for scientific discovery. The system iterates over \emph{learn}, \emph{design}, \emph{experiment}, and \emph{analyze}; each round yields experience that is stored and reused, so that over time the model produces better designs and achieves domain breakthroughs.}
%     \label{fig:teaser}
% \end{figure}

\newpage
\pagestyle{fancy}
\lhead{\rightmark}
\renewcommand{\headrulewidth}{0.7pt}
\setlength{\headsep}{5mm}
\clearpage

\newpage
% Reset footnote numbering for main content
\renewcommand{\thefootnote}{\arabic{footnote}}
\setcounter{footnote}{0}

% \tableofcontents
%% 

\clearpage  % 确保正文从新页开始============================================================
%% Introduction
%% ============================================================
\section{Introduction}
\label{sec:introduction}

Artificial intelligence (AI) advances through many interacting factors; \textbf{data, model architectures, and learning algorithms} are three central research components. Progress in each of these directions depends on repeated cycles of hypothesis generation, implementation, experimentation, and analysis~\citep{ghareeb2025robinmultiagentautomatingscientific}. In practice, however, these cycles are constrained by multidimensional human bottlenecks~\citep{zhang2025positionintelligentsciencelaboratory}: the hypothesis space humans can explore in parallel is severely limited~\citep{liu2025alphagomomentmodelarchitecture}, experimental workflows demand substantial manual effort and frequent intervention~\citep{feng2025aifluidscientistllmpowered}, and the accumulation of insights across iterations often depends on individual experience and intuition, making knowledge difficult to systematically preserve and transfer~\citep{kosmyna2025brainchatgptaccumulationcognitive}. Together, these constraints fundamentally limit the pace and scale of progress in AI development, raising a central question: \emph{can AI accelerate the development of AI itself?}~\citep{Wang2023ScientificDiscovery}

Recent advances in AI capabilities have made this possibility increasingly plausible~\citep{didolkar2024metacognitivecapabilitiesllmsexploration}. The role of AI in scientific discovery has evolved rapidly~\citep{wei2025aiscienceagenticscience}: from specialized systems that solve discrete, well-defined problems such as AlphaFold~\citep{jumper2021highly}, GraphCast~\citep{lam2023graphcastlearningskillfulmediumrange}, and GNoME~\citep{Merchant2023Scaling}, to LLM-based and agentic systems that support broader scientific workflows. Systems such as SciMaster~\citep{chai2025scimastergeneralpurposescientificai} focus on scientific question answering with known answers; ML-Master~\citep{liu2025mlmasteraiforaiintegrationexploration} and MLEvolve~\citep{du2025automlgen} address bounded optimization problems under fixed evaluation criteria; and AI Scientist~\citep{lu2024aiscientist} automates the research publication pipeline rather than tackling open-ended frontier research. AlphaEvolve~\citep{novikov2025alphaevolvecodingagentscientific} takes an important step toward autonomous scientific optimization by iteratively improving candidate solutions through coding agents. Yet the research loops that drive real AI progress remain substantially harder to automate: improving architectures, data pipelines, or training algorithms typically requires modifying large codebases, running costly experiments, interpreting multidimensional outcomes, and sustaining coherent exploration across many rounds. Existing frameworks have not yet demonstrated that AI can operate effectively in this regime in a unified way, nor that it can generate meaningful advances across the \textbf{three foundational pillars of AI development} rather than within a single narrowly scoped setting.

To address this gap, we present \modelname, an agentic framework for \textbf{AI-for-AI} research. The general scientific process follows a principled cycle: researchers collect extensive background literature, formulate informed hypotheses, execute experiments, and distill insights through systematic analysis~\citep{ghareeb2025robinmultiagentautomatingscientific}. Inspired by this workflow, \modelname closes the loop between prior knowledge, hypothesis generation, experimental execution, and iterative refinement through a \emph{learn--design--experiment--analyze} cycle. Two components are central to this design. First, a structured \textbf{cognition base} grounds each round of exploration in accumulated human research literature from the outset, allowing the system to build on domain knowledge rather than search from scratch. Second, a dedicated \textbf{analyzer} translates complex multi-dimensional experimental outcomes into structured, actionable insights that are written back into the experience database for future iterations. Together, these components enable sustained improvement on long-horizon AI research tasks where feedback is expensive, indirect, noisy, and difficult to interpret, substantially improving both the speed and quality of the evolution process.
% \mtt{Perhaps we need some conservative wording.}

Using \modelname, we demonstrate that AI can accelerate multiple parts of its own development stack. \textbf{To our knowledge, this is the first unified demonstration of AI-driven discovery across three central components of AI development: data, architectures, and learning algorithms.} \textbf{(1) Model Architecture}: In neural architecture design, \modelname autonomously generated 1,350 candidates across 1,773 exploration rounds, discovering 105 architectures that surpass the human-designed DeltaNet~\citep{yang2025parallelizinglineartransformersdelta}; its top-performing model achieved a +0.97 point gain, nearly triple the improvement of recent manual SOTA advancements~\citep{dao2024transformersssmsgeneralizedmodels}. \textbf{(2) Data Curation}: In pretraining data curation, evolved strategies produced cleaner training datasets, improving over the original data by 3.96 points on average benchmarks, with particularly strong gains on knowledge-intensive benchmarks such as MMLU~\citep{hendrycks2021measuringmassivemultitasklanguage}, where improvements exceeded 18 points. \textbf{(3) Training Algorithm}: In reinforcement learning algorithm design, the framework derived novel optimization mechanisms with principled mathematical innovations that outperform the competitive GRPO~\citep{Guo_2025} baseline by up to +12.5 points on AMC32, +11.67 points on AIME24, and +5.04 points on OlympiadBench. Together, these results show that under the \modelname framework, AI can accelerate key parts of AI development---from model architecture design to data preparation to learning algorithm development---forming a coherent, end-to-end closed loop for AI self-improvement.

We further validate the effectiveness of \modelname through targeted comparisons and ablation studies. On the circle packing task used as a shared benchmark across evolutionary frameworks, \modelname finds SOTA-level results in as few as 17 rounds, substantially outpacing prior frameworks including OpenEvolve and GEPA. To examine whether AI-designed components provide utility beyond the AI/ML stack, we additionally apply \modelname to drug-target interaction prediction, a biomedical domain distinct from AI development, where the evolved architecture achieves a 6.94-point AUROC improvement in cold-start generalization scenarios. These results provide initial evidence that the AI-for-AI paradigm enabled by \modelname can generalize beyond AI tasks to broader scientific applications.

\section{Preliminary}
\label{sec:related}

% \url{https://arxiv.org/pdf/2602.05688}

% \input{tables/comparison}

% \section{Related Work}

\begin{wrapfigure}{r}{0.5\textwidth}
  \vspace{-\baselineskip}
  \centering
  \includegraphics[width=\linewidth]{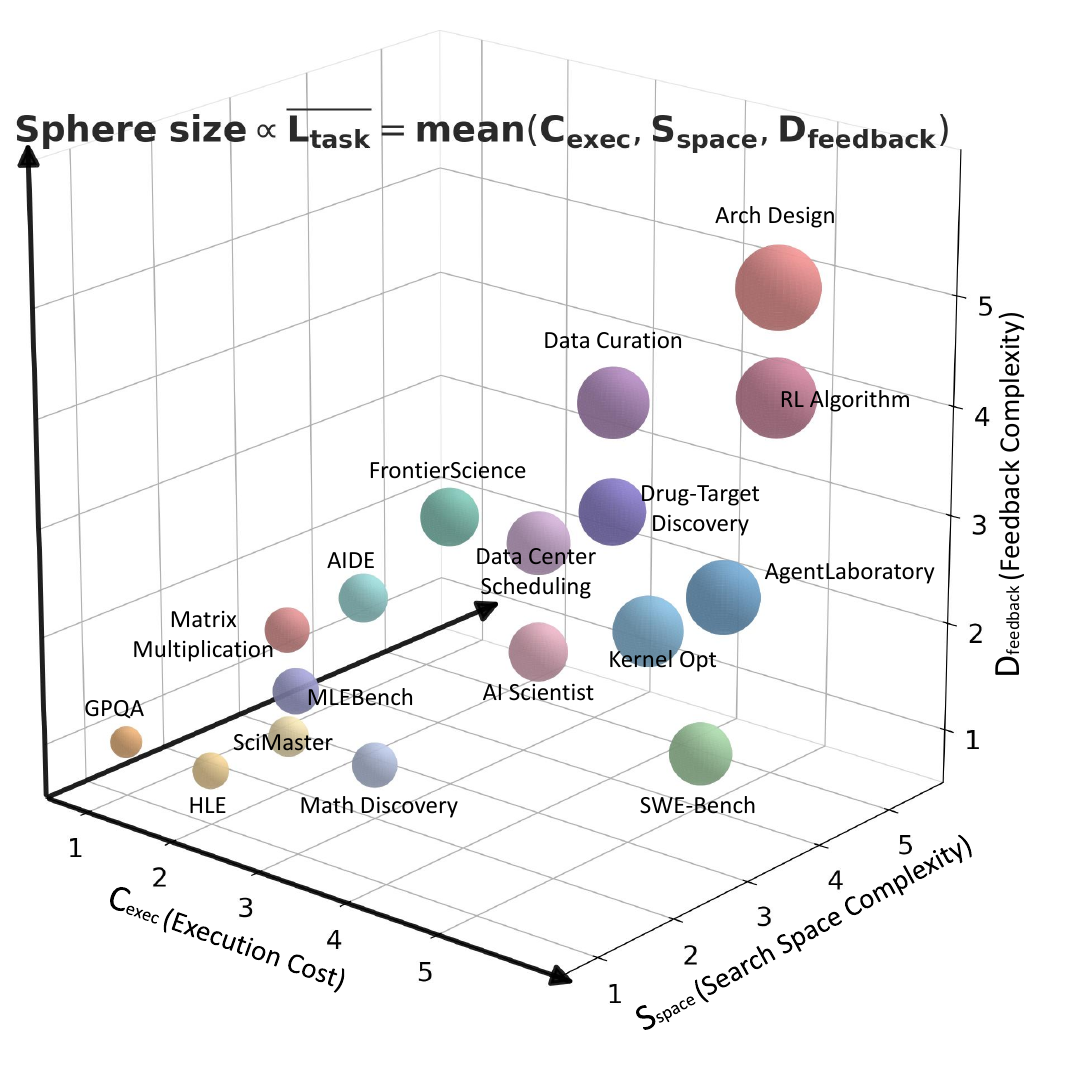}
  \caption{Representative scientific automation settings in the $L_\text{task}=\langle C_\text{exec}, S_\text{space}, D_\text{feedback}\rangle$ space.}
  \label{fig:ltask}
  \vspace{-\baselineskip}
\end{wrapfigure}

% \noindent
To systematically position existing work relative to \modelname, we introduce \textbf{Scientific Task Length ($L_\text{task}$)} as an analytical framework characterizing the intrinsic challenge of autonomous scientific research tasks along three dimensions: (1) Execution Cost ($C_\text{exec}$), which measures the computational resources and engineering complexity required per trial, including the burden of modifying large interdependent codebases and the GPU hours consumed. (2) Search Space Complexity ($S_\text{space}$), which captures the complexity of the solution space the system must navigate, including the openness of the task objective, whether candidate solution boundaries are predefined, and the extent to which meaningful exploration directions must be discovered rather than given. (3) Feedback Complexity ($D_\text{feedback}$), which measures the difficulty of extracting actionable insights from experimental outcomes, reflecting how much the system must synthesize multi-dimensional signals such as loss dynamics, benchmark distributions, and efficiency traces, rather than simply responding to a scalar score. We characterize task complexity as $L_\text{task} = \langle C_\text{exec}, S_\text{space}, D_\text{feedback} \rangle$ and use this lens to survey existing work.

\paragraph{Scientific question answering.}
This class of work involves little to no experimental execution; the task reduces to answering scientific questions against straightforward evaluation criteria, where correctness can be determined without interpreting complex feedback signals or iterative refinement. Idea generation systems~\citep{Wang_2024, hu2024novaiterativeplanningsearch} and automated survey frameworks~\citep{wang2024autosurveylargelanguagemodels} incur virtually zero $C_\text{exec}$. Scientific question answering benchmarks including GPQA~\citep{rein2024gpqa}, HLE~\citep{phan2025humanity}, FrontierScience~\citep{wang2026frontierscienceevaluatingaisability}, and SciMaster~\citep{chai2025scimastergeneralpurposescientificai} similarly operate under simple, univariate evaluation: no multi-dimensional experimental signals need to be synthesized, and no iterative experimental loop is required. All three dimensions of $L_\text{task}$ remain low.

\paragraph{Structured task execution.}
Moving beyond pure reasoning, a number of systems introduce genuine experimental execution under clearly defined objectives. MLE-bench~\citep{chan2024mle-bench} and 
SWE-bench~\citep{jimenez2024swebenchlanguagemodelsresolve} require agents to optimize fixed targets in machine learning competitions or code repair; AIDE~\citep{jiang2025aideaidrivenexplorationspace} performs tree search over code space to optimize user-defined metrics. AI Scientist~\citep{lu2024aiscientist} and 
AgentLaboratory~\citep{schmidgall2025agentlaboratoryusingllm} build more complete end-to-end pipelines, yet ultimately target structured, predefined tasks rather than open scientific discovery. Across this class, tasks follow established patterns with clear success criteria, and the goal is task completion rather than advancing scientific understanding. $C_\text{exec}$ remains modest, and both $S_\text{space}$ and $D_\text{feedback}$ stay bounded: the exploration space is constrained by predefined objectives, and feedback signals, while present, do not require deep synthesis of multi-dimensional experimental outputs to guide meaningful scientific progress.

\paragraph{Lightweight scientific discovery.}
Further along the spectrum, evolutionary search frameworks 
achieve genuine open-ended discovery. 
AlphaEvolve~\citep{novikov2025alphaevolvecodingagentscientific} 
improved Strassen's matrix multiplication algorithm for the 
first time in 56 years, advanced over 50 open mathematical 
problems, and optimized datacenter scheduling and the 
FlashAttention kernel. FunSearch~\citep{FunSearch2023} 
discovered combinatorial optimization algorithms surpassing 
human-designed solutions. Mining Generalizable Activation 
Functions~\citep{vitvitskyi2026mininggeneralizableactivationfunctions} 
uncovered stronger generalizing activation functions through 
evolutionary search. OpenEvolve~\citep{openevolve}, 
GEPA~\citep{agrawal2026gepareflectivepromptevolution}, 
ShinkaEvolve~\citep{lange2025shinkaevolveopenendedsampleefficientprogram}, 
AdaEvolve~\citep{cemri2026adaevolveadaptivellmdriven}, and 
SkyDiscover~\citep{skydiscover2026} further advance this 
paradigm in efficiency, diversity and adaptivity. In the $L_\text{task}$ framework, $S_\text{space}$ and 
$C_\text{exec}$ are both elevated, as objectives are open-ended 
and iterative evaluation is required. Yet each trial remains 
small in scale, with modifications typically localized to a 
single function or short code segment. Feedback is thus 
immediate and direct, keeping $D_\text{feedback}$ low. 

\paragraph{Large-scale scientific exploration.}
At the upper end of the $L_\text{task}$ spectrum lie tasks 
that push all three dimensions to substantially higher levels, 
occupying a region that existing systems have not yet addressed 
(Figure~\ref{fig:ltask}). Neural architecture design, 
pretraining data curation, and training algorithm design are 
foundational to AI progress, and represent three central components 
that \modelname targets. Validating a single candidate 
requires complete model training consuming tens to hundreds of 
GPU hours, often involving deep modifications to large 
interdependent codebases. The exploration space is broad and 
open-ended, spanning diverse design choices with no predefined 
boundaries. Experimental feedback spans multiple benchmarks, 
loss dynamics, and efficiency metrics, all of which must be 
jointly interpreted to guide the next iteration.

These properties impose unique demands on any system that 
attempts to automate such research. Each experimental trial is 
costly and opportunities for iteration are limited, meaning 
the system cannot afford to explore blindly. Prior domain 
knowledge must therefore be incorporated from the outset to 
steer exploration toward promising directions, motivating the 
\textbf{Cognition Base} in \modelname{}. At the same time, 
the richness of experimental feedback calls for dedicated 
interpretation: raw signals across benchmarks and training 
dynamics must be distilled into actionable insights before 
the next iteration can proceed, motivating the structured 
\textbf{Analyzer}. Together, these components reflect a 
key distinction between \modelname{} and existing evolutionary 
frameworks: while prior work evolves candidate solutions, 
\modelname{} evolves cognition itself. Accumulated experience 
and distilled insights are continuously stored and retrieved 
to inform future exploration, ensuring that the system grows 
not only in the quality of its solutions but in its capacity 
to reason about where to search next.

\section{ASI-Evolve}
\label{sec:method}

\begin{figure}[t]
    \centering
    \includegraphics[width=\linewidth]{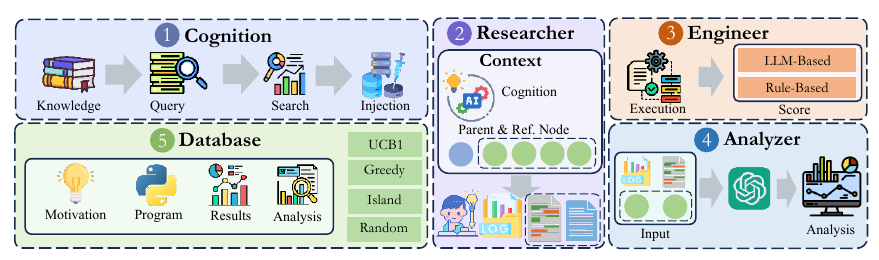}
    \caption{\modelname{} pipeline: in each round, the system samples context nodes from database, retrieves relevant cognition items via embedding search, generates a new candidate program, runs an experiment-specific evaluation script under timeouts, and summarizes results into an analysis report that is stored back into the database for future rounds.}
    \label{fig:pipeline}
\end{figure}

\modelname{} is implemmented as an end-to-end experimental evolution pipeline. As illustrated in Figure~\ref{fig:pipeline}, each iteration proceeds through four stages:  (i) learn relevant knowledge and historical experience respectively from cognition storage and database, (iii) design the next candidate program, (iv) executes an experiment to obtain evaluation signals, and (v) analyzes outcomes into reusable, human-readable lessons.  Below we describe the four corresponding modules in the actual system.

We view each evolution round $t$ as searching over a program space $\mathcal{P}$ (code artifacts that implement a solution). The system maintains: (1) a database $\mathcal{D}$ of past nodes (motivation, code, results, analysis, score, and metadata), and (2) a cognition store $C$ of task-relevant textual items indexed by embeddings. A new candidate program is generated conditioned on sampled context nodes $S_t \sim \mathrm{Sample}(\mathcal{D})$ and retrieved cognition items $R_t = \mathrm{Retrieve}(C;\, S_t)$,
\[
p_t \sim P(p \mid S_t, R_t),
\]
and is evaluated by an external, experiment-specific procedure that produces structured metrics with a primary scalar score. The resulting node is then appended to $\mathcal{D}$ for subsequent sampling. 
% \mtt{a little confusing}

\subsection{Researcher}
\label{subsec:researcher}

The Researcher generates the next candidate program given the task description, sampled context nodes, and retrieved cognition items. Each round it begins by sampling $n$ nodes from the database, and then retrieving a small set of cognition items by semantic search over the sampled nodes' analyses or motivations to provide additional priors.

Conditioned on this context, the Researcher uses an LLM to produce a complete program together with a natural-language motivation, which are stored together as a new node for subsequent rounds. In addition to full-code generation, the system also supports an optional \emph{diff-based} editing mode, where the model proposes localized modifications over a parent program; this incremental style is particularly helpful when evolving larger codebases over many rounds.

% The researcher is responsible for proposing new designs given historical experience. At the start of each round, the system samples parent architectures from the database and retrieves relevant past experiment records as reference. When context length is excessive, an LLM can summarize lengthy descriptions (e.g., motivation, implementation, analysis) from the history. ASI-Evolve uses a single unified agent for both motivation generation and code implementation, so that design rationale and implementation details stay consistent. Before submitting an experiment, the system uses embedding-based retrieval over top-$k$ similar motivations to assess novelty, reducing the chance of redundant experiments.

\subsection{Engineer}
\label{subsec:engineer}

The Engineer executes the candidate program in the actual experiment environment and produces the quantitative evaluation signal used for evolution. Given a generated program, it invokes a user-specified evaluation procedure that runs the experiment end-to-end and returns structured metrics, including a primary scalar score that serves as the fitness signal.

To better handle long-horizon tasks, the Engineer supports early rejection via configurable wall-clock limits and lightweight quick tests, improving efficiency by filtering flawed candidates before expensive runs. It can also optionally invoke an LLM-based judge to cover aspects of candidate quality that are difficult to capture with rule-based metrics alone, combining its score with the primary metric.

% The engineer executes experiments and evaluation. Given task difficulty and high run cost, we first perform heuristic static checks (e.g., complexity, mask pattern); designs that fail are returned for revision to avoid wasted compute. We then run experiments in the real code environment; when implementation errors occur, the error log is returned to the agent for self-debug and code fix, so that promising designs are not abandoned due to minor bugs. During training, the system monitors loss curves, compile time, and related metrics in real time and terminates abnormal runs (e.g., unusually slow training, unreasonably low loss) to save resources. Besides quantitative metrics, we optionally use LLM-as-a-Judge to score results qualitatively along dimensions such as complexity, efficiency, and innovativeness; these scores are combined with benchmark performance to form a final score that guides evolution.

\subsection{Analyzer}
\label{subsec:analyzer}
In our setting, the primary feedback used for selection is a scalar score produced by a task-specific evaluation, but the same run also yields rich auxiliary signals—multiple metrics, feature importances, training logs, and execution traces—that are useful for diagnosis yet too verbose to feed directly into subsequent rounds. This is particularly pronounced in the complex, large-scale tasks we target, where a single experiment may generate extensive logs spanning training dynamics, benchmark breakdowns, and efficiency traces. The Analyzer is designed to handle this asymmetry: it receives the current program together with the full experimental output—including raw logs and detailed metrics—and distills them into a compact, decision-oriented report. This full exposure allows the Analyzer to perform thorough causal analysis; the resulting report will be persisted in the database and used for retrieval in subsequent rounds, keeping context length manageable without sacrificing analytical depth.

% Concretely, the Analyzer is an LLM agent that optionally also receives a reference node from the sampled context. It produces a concise analysis that highlights what changed, what likely drove the score change, and what actionable next modifications to try. This analysis is persisted in the database and becomes the primary textual signal used for retrieval in subsequent rounds.

\subsection{Cognition}
\label{subsec:cognition}

For long-horizon research tasks, exploration from scratch offers a larger hypothesis space but incurs substantial resource and time cost. We therefore introduce a \textbf{cognition base} that encodes \textbf{human prior knowledge}---task-relevant heuristics, known pitfalls, and design principles drawn from domain literature and prior runs---so that the system can be steered toward promising directions and iterate efficiently rather than rediscovering well-documented failure modes. In each round, after sampling context nodes from the database, the pipeline uses the sampled nodes' information as queries to \emph{retrieve} a small set of similar cognition entries via embedding-based semantic search; these entries are then injected into the Researcher's context to guide hypothesis generation. Our experiments (see~\S\ref{sec:ablation}) show that equipping the loop with this cognition base significantly improves cold-start climb speed and iteration efficiency, without limiting long-term exploration capability.

\subsection{Database}
\label{subsec:database}

The database is the system's persistent memory: it stores the outcome of each evolution round and supplies the \emph{sampled} nodes that form the Researcher's context. Whereas the cognition base provides fast-start prior knowledge, the historical nodes in the database convey \textbf{task-specific} information and become the dominant information source as evolution progresses, supporting sustained improvement beyond the initial climb. Each evolution step produces a \emph{node} that stores: (i) the Researcher motivation, (ii) the generated program, (iii) structured results from the evaluation script, (iv) analysis report, and (v) auxiliary metadata such as runtime and success flag. For sampling, to support comparison and flexible deployment we encapsulate multiple policies behind a unified interface---UCB1, random, greedy, and MAP-Elites island algorithm. Our ablation studies (see~\S\ref{sec:ablation}) show that the choice of sampling algorithm strongly affects \emph{sustained} improvement: unlike the cognition base, which primarily accelerates cold-start climb, different sampling strategies yield clearly distinct evolution trajectories and are therefore critical for tuning the system to a given task.

\section{Main Tasks}
\label{sec:main-tasks}

We apply \modelname{} to three central components of AI development---model architecture design, training data preparation, and training algorithm design---covering key parts of the AI research pipeline from model structure to data to training. Each task shares a set of characteristics that make autonomous research particularly challenging: limited available prior knowledge specific to the task, long iteration cycles, substantial implementation complexity, and experimental feedback that is indirect, multi-dimensional, and difficult to interpret. These properties collectively define a regime that poses significant challenges to automated scientific research, making them a demanding testbed for evaluating \modelname{}'s capability to conduct autonomous AI research at scale.

\input{exp_architecture.tex}
\input{exp_data.tex}

\input{exp_rl.tex}

%% ============================================================
%% Ablation Study
%% ============================================================
\input{ablation.tex}
\input{exp_medical.tex}
\section{Conclusion}
\label{sec:conclusion}

In this paper, we presented \modelname, an agentic evolution framework that enables AI to carry out end-to-end autonomous scientific research. Through controlled comparisons against existing evolutionary baselines and systematic ablation studies, we verified that the framework design is effective: equipped with a structured cognition base and a dedicated analyzer, the system achieves rapid cold-start and sustains continuous improvement, reliably reaching SOTA-level results.

We further explored whether AI can accelerate its own research pipeline across each stage of the scientific process. The closed \emph{learn--design--experiment--analyze} loop enables efficient self-improvement, and we demonstrate breakthroughs across three central components of AI development---model architecture, training data, and training algorithms---each posing substantial challenges in terms of implementation complexity, iteration cost, and indirect feedback. Beyond the core AI pipeline, our drug-target interaction experiment demonstrates that model designs discovered through AI-driven research can be effectively deployed in real-world tasks, showing that AI-optimized solutions carry genuine scientific value.

Looking ahead, the scope of AI self-acceleration extends beyond individual models to the full AI development stack---architecture, data, algorithms, and infrastructure yet to be explored. As agentic systems take on more of the implementation and iteration work, human scientists can shift from being the executors of solutions to the definers of problems---concentrating their expertise on the questions that matter most and leaving the expansive search through hypothesis spaces to AI. We expect this paradigm to drive not only the self-improvement of individual models, but the self-evolution of the entire AI field.

\bibliographystyle{acl_natbib}
\bibliography{bib}
\clearpage

\appendix
\input{appendix/exp_ablation.tex}
\end{document}

%% file: setup.tex
\usepackage{authblk}
\usepackage[utf8]{inputenc}
\usepackage[T1]{fontenc}
\usepackage{main} % ACL style file
\usepackage{microtype}
\usepackage{times}
\usepackage{latexsym}
\usepackage{algorithm}
\usepackage{algorithmic}
\usepackage{graphicx}
\usepackage{wrapfig}
\usepackage{xcolor}
\definecolor{deepgreen}{RGB}{34,139,34}

\usepackage{adjustbox}
\usepackage{booktabs}
\usepackage{amsmath}
\usepackage{amssymb}
\usepackage{amsfonts}
\usepackage{mathtools}
\usepackage{bm}

\usepackage{graphicx}
\usepackage{subcaption}
\usepackage{float}
\usepackage{wrapfig}

\usepackage{booktabs}
\usepackage{array}
\usepackage{multirow}
\usepackage{multicol}
\usepackage{makecell}
\usepackage{longtable}
\usepackage{tabularx}
\usepackage{colortbl}
\usepackage{arydshln}

\usepackage{color}
\usepackage{xcolor}
\definecolor{mydarkblue}{rgb}{0,0.08,0.45}
\definecolor{wkblue}{rgb}{0.2, 0.3, 0.6}
\definecolor{meta-color}{rgb}{0.5, 0.5, 0.5}
\definecolor{bgblue}{RGB}{245,243,253}
\definecolor{ttblue}{RGB}{91,194,224}
\definecolor{mybrown}{RGB}{128,64,0}
\definecolor{titlecolor}{HTML}{4c9cff}
\definecolor{coolblue3}{rgb}{0.91, 0.94, 0.98}
\definecolor{myblue}{rgb}{0.9, 0.1, 0.94}
\definecolor{mygreen}{rgb}{0.64, 0.56, 0.88}
\definecolor{myyellow}{rgb}{0.68, 0.6, 0.1}
\definecolor{fancygreen}{rgb}{0.33, 0.68, 0.20}
\definecolor{salmon}{rgb}{0.94, 0.52, 0.49}
\definecolor{tablegreen}{rgb}{0.82, 0.94, 0.75}
\definecolor{tableblue}{rgb}{0.81, 0.90, 0.94}
\definecolor{tablered}{rgb}{0.97, 0.85, 0.85}
\definecolor{tableorange}{rgb}{0.96, 0.85, 0.81}

\usepackage{enumitem}
\usepackage{footnote}
\usepackage{lipsum}
\usepackage{setspace}

\usepackage{geometry}
\usepackage{fancyhdr}
\usepackage{lscape}
\usepackage{rotating}

\usepackage{algorithm}
\usepackage{algorithmic}

\usepackage[most]{tcolorbox}
\usepackage[framemethod=tikz]{mdframed}
\usepackage{awesomebox}

\usepackage{natbib}
\usepackage{url}
\usepackage[colorlinks=true,linkcolor=mydarkblue,citecolor=mydarkblue,filecolor=mydarkblue,urlcolor=mydarkblue]{hyperref}

\usepackage{bbding}
\usepackage{imakeidx}
\makeindex
\usepackage[edges]{forest}
\usepackage[normalem]{ulem}
\usepackage{fontawesome5}
\usepackage{blindtext}
\usepackage{pgfplots}
\usepackage{pgfplotstable}

\usepackage{listings}
\usepackage{listingsutf8}
\newcommand\JSONnumbervaluestyle{\color{blue}}
\newcommand\JSONstringvaluestyle{\color{red}}

\newif\ifcolonfoundonthisline

\makeatletter
\lstdefinestyle{json}
{
  showstringspaces    = false,
  keywords            = {false,true},
  alsoletter          = 0123456789.,
  morestring          = [s]{"}{"},
  stringstyle         = \ifcolonfoundonthisline\JSONstringvaluestyle\fi,
  MoreSelectCharTable =%
    \lst@DefSaveDef{`:}\colon@json{\processColon@json},
  basicstyle          = \ttfamily,
  keywordstyle        = \ttfamily\bfseries,
}

\newcommand\processColon@json{%
  \colon@json%
  \ifnum\lst@mode=\lst@Pmode%
    \global\colonfoundonthislinetrue%
  \fi
}

\lst@AddToHook{Output}{%
  \ifcolonfoundonthisline%
    \ifnum\lst@mode=\lst@Pmode%
      \def\lst@thestyle{\JSONnumbervaluestyle}%
    \fi
  \fi
  \lsthk@DetectKeywords%
}

\lst@AddToHook{EOL}%
  {\global\colonfoundonthislinefalse}
\makeatother

\newtcolorbox{myboxi}[1][]{
  breakable,
  title=#1,
  colback=red!5,
  colbacktitle=red!5,
  coltitle=black,
  fonttitle=\bfseries,
  bottomrule=0pt,
  toprule=0pt,
  leftrule=2pt,
  rightrule=2pt,
  titlerule=0pt,
  arc=0pt,
  outer arc=0pt,
  colframe=red,
}

\newtcolorbox{myboxnote}[1][]{
  breakable,
  title=#1,
  colback=orange!0,
  colbacktitle=orange!0,
  coltitle=black,
  fonttitle=\bfseries,
  bottomrule=0pt,
  toprule=0pt,
  leftrule=2pt,
  rightrule=2pt,
  titlerule=0pt,
  arc=0pt,
  outer arc=0pt,
  colframe=orange,
}

\newtcolorbox{myboxii}[1][]{
  breakable,
  freelance,
  title=#1,
  colback=white,
  colbacktitle=white,
  coltitle=black,
  fonttitle=\bfseries,
  bottomrule=0pt,
  boxrule=0pt,
  colframe=white,
  overlay unbroken and first={
  \draw[red!75!black,line width=3pt]
    ([xshift=5pt]frame.north west) --
    (frame.north west) --
    (frame.south west);
  \draw[red!75!black,line width=3pt]
    ([xshift=-5pt]frame.north east) --
    (frame.north east) --
    (frame.south east);
  },
  overlay unbroken app={
  \draw[red!75!black,line width=3pt,line cap=rect]
    (frame.south west) --
    ([xshift=5pt]frame.south west);
  \draw[red!75!black,line width=3pt,line cap=rect]
    (frame.south east) --
    ([xshift=-5pt]frame.south east);
  },
  overlay middle and last={
  \draw[red!75!black,line width=3pt]
    (frame.north west) --
    (frame.south west);
  \draw[red!75!black,line width=3pt]
    (frame.north east) --
    (frame.south east);
  },
  overlay last app={
  \draw[red!75!black,line width=3pt,line cap=rect]
    (frame.south west) --
    ([xshift=5pt]frame.south west);
  \draw[red!75!black,line width=3pt,line cap=rect]
    (frame.south east) --
    ([xshift=-5pt]frame.south east);
  },
}

\mdfdefinestyle{mystyle}{%
  rightline=true,
  innerleftmargin=10,
  innerrightmargin=10,
  outerlinewidth=3pt,
  topline=false,
  rightline=true,
  bottomline=false,
  skipabove=\topsep,
  skipbelow=\topsep
}

\usepackage{pifont}
\DeclareCaptionFont{black}{\color{black}}

\newenvironment{itemize*}%
 {\leftmargini=10pt\begin{itemize}%
  \setlength{\itemsep}{0pt}%
  \setlength{\parskip}{0pt}%
  }%
 {\end{itemize}}
\newenvironment{enumerate*}%
 {\begin{enumerate}%
  \setlength{\itemsep}{0pt}%
  \setlength{\parskip}{0pt}}%
 {\end{enumerate}}

\usepackage{etoolbox}
\newcounter{bibcount}
\makeatletter
\patchcmd{\@lbibitem}{\item[}{\item[\hfil\stepcounter{bibcount}{[\thebibcount]}}{}{}
\renewcommand\NAT@bibsetup%
  [1]{\setlength{\leftmargin}{\bibhang}\setlength{\itemindent}{-\parindent}%
      \setlength{\itemsep}{\bibsep}\setlength{\parsep}{\z@}}
\makeatother

\definecolor{myblue}{rgb}{0.9, 0.1, 0.94}
\definecolor{mygreen}{rgb}{0.64, 0.56, 0.88}
\definecolor{myyellow}{rgb}{0.68, 0.6, 0.1}
\definecolor{fancygreen}{rgb}{0.33, 0.68, 0.20}
\definecolor{salmon}{rgb}{0.94, 0.52, 0.49}
\definecolor{tablegreen}{rgb}{0.82, 0.94, 0.75}
\definecolor{tableblue}{rgb}{0.81, 0.90, 0.94}
\definecolor{tablered}{rgb}{0.97, 0.85, 0.85}
\definecolor{tableorange}{rgb}{0.96, 0.85, 0.81}

\newcommand{\modelname}{\textsc{Asi-Evolve}\xspace}

%% file: exp_architecture.tex
% exp_architecture.tex - Model Architecture Experiment
% This file contains the Model Architecture experiment section

\subsection{Scenario 1:  Model Architecture Design}
\label{subsec:arch}

\paragraph{Task Formulation}
Model architecture is a foundational component of AI systems, determining the capacity to model complex patterns, computational efficiency, and generalization. In this task, we focus on designing efficient sequence models through linear attention mechanisms. The quadratic complexity of standard Transformer attention ($O(N^2)$) has motivated extensive research into sub-quadratic alternatives---including DeltaNet~\cite{yang2025parallelizinglineartransformersdelta}, Gated DeltaNet~\cite{yang2024gateddeltanet}, Mamba~\cite{dao2024transformersssmsgeneralizedmodels}, and RWKV~\cite{peng2023rwkv}---which achieve $O(N)$ complexity by decomposing attention computations or maintaining compressed memory states. Despite this progress, improving efficiency while preserving modeling capacity remains challenging, and the design space remains vast and under-explored. Using DeltaNet~\cite{yang2025parallelizinglineartransformersdelta} as the baseline, the task requires the AI system to design novel attention layers with sub-quadratic complexity, employ chunk-wise computation patterns for efficient parallel training, and produce complete runnable implementations integrated into an existing large codebase.
% This encapsulates the key difficulties of agentic scientific discovery: balancing mathematical complexity-capacity trade-offs, modifying thousands of lines of interdependent code, satisfying hard constraints (complexity bounds, chunk-wise patterns, causal masking) alongside soft objectives (performance, efficiency), and conducting quality assessment through costly GPU-hour experiments with multi-dimensional feedback.

\paragraph{Methodology}
We initialize the cognition repository with approximately 150 entries extracted from 100 papers on linear attention, state space models, and efficient transformers, providing the system with domain priors from the outset. The database uses a periodically refreshed candidate pool that retains the top 50 highest-scoring nodes; each round samples its root architecture from the top 10 and draws reference context from the broader top 50. Two characteristics of this task further motivate targeted engineering adaptations: each evaluation costs hours of GPU training, and the design space contains numerous hard constraints that are easy to violate. To improve runtime efficiency and constraint satisfaction, we introduce three mechanisms. A \textit{static check agent} intercepts each generated design before training, verifying complexity bounds, chunk-wise structure, and causal mask correctness. A \textit{debug agent} handles runtime implementation errors by inspecting error traces and attempting targeted fixes. A \textit{novelty check} filters duplicate proposals via motivation similarity, encouraging genuine exploration.

\input{tables/architecture/id_ood_1p3b.tex}

We adopt a multi-stage evaluation strategy to balance exploration efficiency with result reliability. In the \textit{exploration phase}, small models ($\sim$20M parameters: 8 layers, hidden dimension 256) are trained for 2000 steps on 1B tokens and evaluated on 10 core benchmarks with 500 samples each. Candidates are scored via a composite fitness that combines quantitative metrics from loss and benchmark scores after sigmoid normalization with LLM-as-a-Judge qualitative scores for code complexity, efficiency, and innovativeness; only architectures exceeding the baseline on both dimensions advance. In the \textit{verification phase}, promising candidates are scaled to $\sim$340M parameters (24 layers, hidden dimension 1024) and trained on 1B tokens to verify that their gains persist under scaling. We also conduct additional validity checks, including causality tests to ensure that attention masks correctly prevent future information leakage. The top architectures then undergo \textit{large-scale validation} at $\sim$1.3B parameters (24 layers, hidden dimension 2048), trained on 100B tokens, with evaluation expanded to 16 benchmarks including 6 held-out OOD test sets covering mathematics, code understanding, and multilingual tasks.

\paragraph{Results}

Over 1773 exploration rounds, \textbf{105} architectures surpassed the DeltaNet baseline in the verification phase. We selected 5 representative architectures spanning diverse design philosophies for large-scale validation. Table~\ref{tab:id_ood_1p3b} presents their full performance across development and generalization benchmarks. On development benchmarks, these architectures achieve up to 57.28\% average accuracy compared to DeltaNet's 55.76\%; on generalization benchmarks, they reach up to 45.40\% versus DeltaNet's 44.74\%, confirming that gains transfer beyond the training distribution. Our best model achieves nearly \textbf{3$\times$} the gain of the current human-designed SOTA (Mamba2's +0.34 points over DeltaNet~\cite{dao2024transformersssmsgeneralizedmodels}). This demonstrates that AI-driven evolution can discover architectures significantly outperforming human expert designs even in this high-saturation regime.

\paragraph{Analysis}
Analysis of the top 5 architectures reveals a consistent theme: moving beyond fixed allocation schemes toward adaptive, multi-scale routing that dynamically adjusts computational budget based on input content. \textbf{PathGateFusionNet} introduces hierarchical routing where a first-stage gate allocates budget between local and contextual processing, and a second stage distributes the contextual budget across short-range, long-range, and delta-rule update paths. \textbf{ContentSharpRouter} implements content-aware routing with learnable temperature parameters that prevent premature commitment to single pathways. \textbf{FusionGatedFIRNet} replaces softmax routing with independent sigmoid gates, allowing simultaneous activation of local and global paths alongside per-head retention parameters for the delta-rule memory path. \textbf{HierGateNet} employs two-stage gating with dynamic learnable floor values ensuring critical paths---especially the delta-path for long-range reasoning---never fully collapse. \textbf{AdaMultiPathGateNet} achieves token-level control via a unified BalancedSparseGate combining global, per-head, and per-token logits with entropy penalties preventing mode collapse. These architectures collectively demonstrate that principled adaptive routing, rather than fixed structural choices, is the key lever for improving upon the DeltaNet baseline.

We conduct multi-faceted analyses to understand why the framework works and how fitness guides the evolutionary trajectory. To study this guidance mechanism, we track the fitness and performance curves throughout the search. As expected, sigmoid normalization progressively compresses rule-based score differences in later rounds, giving high-scoring nodes relatively uniform sampling opportunities rather than allowing benchmark leaders to dominate. Correspondingly, 78\% of high-performing architectures discovered in the second half are improvements built on designs found before round 900, indicating that \modelname{} follows the intended fitness design: benchmark-driven exploration in the early stage, followed by broader subjective refinement in later stages. To assess the role of cognition, we further compare design provenance across all 1773 architectures and the 105 SOTA architectures. Across the full population, 51.7\% derive from the cognition base, 38.2\% from accumulated experience, and 10.1\% are novel; among SOTA architectures, the share from experience rises to 44.8\% while novelty drops to 6.6\%. This shift suggests that domain priors effectively accelerate cold-start search, while useful patterns are progressively distilled from the system's own trials as evolution proceeds. One notable limitation concerns computational efficiency: because the system operates at the level of attention mechanism design rather than low-level kernel implementation, it cannot directly produce hardware-optimized CUDA kernels. Although LLM-as-a-Judge scores penalize computationally expensive designs, this qualitative signal cannot guarantee that discovered architectures will match the wall-clock efficiency of top human-engineered implementations after full optimization.

%% file: tables/architecture/id_ood_1p3b.tex
\begin{table*}[!ht]
  \centering
  \footnotesize
\renewcommand{\arraystretch}{1.15}
\resizebox{0.92\linewidth}{!}{%
\begin{tabular}{lcccccccc}
    \toprule
    & \multicolumn{3}{c}{\textbf{Human Discovered \faUserGraduate}} & \multicolumn{5}{c}{\textbf{AI Discovered \faRobot}} \\
    \cmidrule(lr){2-4}\cmidrule(l){5-9}
    \textbf{Benchmarks} & \textbf{DeltaNet} & \textbf{\makecell{Gated-\\DeltaNet}} & \textbf{Mamba2} & \textbf{PG} & \textbf{C} & \textbf{FG} & \textbf{H} & \textbf{AM} \\
    \midrule
    \rowcolor{gray!15}\multicolumn{9}{c}{\emph{Development}} \\
    \midrule
    Wiki ppl$\downarrow$ & 17.00 & 16.84 & 16.66 & 16.18 & \underline{16.05} & \textbf{15.77} & 16.65 & 16.26 \\
   LMB ppl$\downarrow$ & 13.63 & 13.31 & 13.33 & \underline{12.62} & 13.45 & \textbf{12.34} & 13.06 & 13.75 \\
   LMB & 45.47 & 46.26 & 46.24 & \textbf{47.60} & 46.13 & \underline{47.53} & 46.56 & 45.04 \\
   PIQA & 73.12 & 74.10 & 73.78 & 72.91 & \underline{74.37} & 72.91 & \textbf{74.37} & 74.10 \\
   Hella & 56.29 & 57.55 & \textbf{58.58} & 56.99 & 57.00 & \underline{58.47} & 56.85 & 57.17 \\
   Wino & 55.88 & 58.01 & \underline{58.48} & 57.22 & 57.85 & \textbf{60.14} & 57.38 & 57.62 \\
   ARC-e & 73.40 & 72.14 & 72.98 & 73.06 & 72.05 & \underline{74.28} & 73.11 & \textbf{74.28} \\
   ARC-c & \textbf{40.61} & 36.95 & 39.33 & \underline{40.36} & 39.76 & 40.02 & 39.33 & 39.33 \\
   SIQA & 40.74 & 41.71 & 41.81 & \underline{42.37} & 41.81 & \textbf{42.78} & 42.07 & 42.07 \\
   BoolQ & 60.58 & 53.98 & 60.52 & 62.45 & \underline{62.51} & 62.11 & \textbf{63.03} & 56.27 \\
    \midrule
    \rowcolor{gray!15}\multicolumn{9}{c}{\emph{Generalization}} \\
    \midrule
   RACE & 34.45 & 33.78 & 32.15 & 35.22 & 34.55 & \textbf{35.60} & \underline{35.22} & 35.02 \\
   BBQ & 29.53 & 29.75 & 29.43 & 29.95 & 30.55 & \textbf{31.46} & \underline{30.88} & 30.27 \\
   MetaBench & 26.97 & 28.67 & 27.70 & 25.64 & \textbf{29.55} & 26.79 & \underline{29.38} & 28.98 \\
   QA4MRE & \textbf{40.00} & 35.00 & 39.17 & 39.17 & \underline{39.17} & 38.33 & 38.33 & 38.33 \\
   SCIQ & 89.80 & \underline{90.30} & 90.30 & 89.60 & 89.50 & 89.20 & \textbf{90.40} & 89.20 \\
   SWAG & 47.69 & 48.17 & \textbf{48.88} & 48.22 & 47.80 & \underline{48.57} & 48.18 & 47.80 \\
    \midrule
    \rowcolor{gray!15}\multicolumn{9}{c}{\emph{Averages}} \\
    \midrule
   Dev. Avg & 55.76 & 55.09 & 56.47 & \underline{56.62} & 56.44 & \textbf{57.28} & 56.59 & 55.74 \\
   Gen. Avg & 44.74 & 44.28 & 44.61 & 44.63 & \underline{45.19} & 44.99 & \textbf{45.40} & 44.93 \\
    Overall Avg & 51.04 & 50.46 & 51.38 & 51.48 & 51.61 & \textbf{52.01} & \underline{51.79} & 51.11\\
    \bottomrule
  \end{tabular}}
\caption{Top block: 10 development benchmarks, used in our exploration stage; middle block: 6 generalization benchmarks for out-of-distribution testing. \textbf{Bold} indicates the best result and \underline{underline} is the suboptimal one. Model abbreviations are as follows: PG = PathGate-FusionNet, C = Content-SharpRouter, FG = FusionGated-FIRNet, H = Hier-GateNet, and AM = AdaMulti-PathGateNet.}
  \label{tab:id_ood_1p3b}
\end{table*}

%% file: exp_data.tex
% exp_data.tex - Pretraining Data Curation Experiment
% This file contains the Pretraining Data Curation experiment section

\subsection{Scenario 2: Pretraining Data Curation}
\label{subsec:data}

\paragraph{Task Formulation}
In this task, the Evolve system must design category-specific curation strategies that improve pretraining data quality. Strategy design is inherently difficult: the strategy space is vast and discrete, encompassing choices of which operations to apply, how to specify decision criteria, and which quality issues to prioritize, with no clear mapping from design choices to effectiveness. For each category, experts must examine data samples to identify issues, explore this combinatorial space to formulate candidate strategies, write detailed specifications, validate results, and iteratively refine, a process requiring significant effort per category. This challenge scales with corpus heterogeneity: modern pretraining corpora comprise hundreds of categories spanning domains, content types and quality levels, each demanding independent strategy design. The AI must automate this process by observing data to identify issues, generating candidate strategies, evaluating effectiveness through diagnostic feedback, and refining iteratively to discover effective approaches across all categories.

\paragraph{Methodology}
We apply the ASI-Evolve framework to the pretraining data curation task. The cognition repository is initialized by examining sampled data from each category, storing identified quality issues such as HTML artifacts, incomplete fragments, formatting inconsistencies, and domain-specific noise patterns. In each iteration, the Researcher retrieves relevant quality issues from the cognition repository and generates candidate curation strategies. The Engineer executes these strategies on 500 sampled documents, applying the specified operations to produce cleaned versions. The Analyzer evaluate 50 randomly selected (original, cleaned) pairs, scoring each on a 1-10 scale. The Analyzer also provides diagnostic feedback on coverage (which identified issues were addressed) and executability (instruction clarity and consistency). The Database maintains complete records of all designed strategies with their scores and diagnostic analyses. Newly discovered quality issues during evaluation are added back to the cognition repository. These feedback mechanisms guide strategy refinement in subsequent iterations.

\paragraph{Results}

\begin{table*}[htbp]
\centering
\setlength{\tabcolsep}{4pt}
\renewcommand{\arraystretch}{0.85}
\small
\begin{tabular}{lcccccc}
\toprule
& \multicolumn{4}{c}{\textbf{Human Discovered \faUserGraduate}} & \multicolumn{2}{c}{\textbf{AI  Discovered \faRobot}} \\
\cmidrule(lr){2-5}\cmidrule(l){6-7}
\textbf{Benchmark} & \textbf{Fineweb-Edu} & \textbf{Ultra-Fineweb} & \textbf{DCLM} & \textbf{Nemotron-CC} & $\textbf{Nemotron-CC}_{\textbf{ASI}}$ & $\textbf{Nemotron-CC}_{\textbf{ASI+}}$ \\
\midrule
BBH        & 3.01  & 7.42  & 24.16 & \textbf{26.82} & \underline{26.69} & 26.16 \\
ARC-E      & 73.39 & 73.96 & 75.13 & 74.94 & \underline{77.55} & \textbf{78.59} \\
ARC-C      & 43.45 & 43.77 & 45.02 & 43.52 & \underline{48.41} & \textbf{49.32} \\
MMLU       & 28.38 & 25.53 & 28.54 & 27.49 & \underline{32.55} & \textbf{46.13} \\
AGIEval    & 16.96 & 17.72 & 17.90 & 18.15 & \textbf{18.30} & \underline{18.21} \\
HellaSwag  & 64.33 & \underline{65.32} & \textbf{70.39} & \underline{65.32} & 64.36 & 62.21 \\
TriviaQA   & 0.67  & 0.42  & \textbf{42.85} & 25.33 & \underline{26.96} & 26.65 \\
RACE       & 35.43 & 34.28 & \textbf{36.08} & 35.04 & \underline{35.63} & 34.28 \\
DROP       & 6.78  & 7.78  & \textbf{24.31} & \underline{19.57} & 19.48 & 18.49 \\
WinoGrande & \underline{61.02} & 60.93 & \textbf{64.99} & 57.96 & 59.89 & 58.09 \\
PIQA       & 75.80 & 75.63 & \textbf{77.93} & 76.79 & \underline{76.80} & 76.15 \\
CSQA       & 19.54 & 19.90 & 20.16 & 20.31 & \underline{20.61} & \textbf{39.12} \\
SIQA       & \underline{45.02} & 43.43 & \textbf{47.51} & 44.36 & 43.58 & 43.57 \\
OpenBookQA & 39.92 & 39.84 & \textbf{43.36} & 39.80 & \underline{42.20} & 41.44 \\
GPQA       & 24.51 & 23.04 & \underline{25.67} & 24.37 & 23.93 & \textbf{27.10} \\
MedQA      & 26.36 & 24.84 & 24.88 & \underline{26.77} & 26.11 & \textbf{40.25} \\
MedMCQA    & 25.80 & 24.92 & 28.15 & 28.86 & \underline{30.28} & \textbf{40.97} \\
PubMedQA   & 67.04 & 64.44 & 66.56 & \underline{67.68} & \textbf{68.32} & \underline{67.68} \\
Avg        & 36.52 & 36.29 & \underline{42.42} & 40.17 & 41.20 & \textbf{44.13} \\
\bottomrule
\end{tabular}
\caption{Benchmark comparison of models trained on different pretraining corpora and curation strategies.
All models are 3B parameters trained on 500B tokens.
Columns are grouped into Human baselines and ASI-Evolve discovered datasets.}
\label{tab:combined}
\end{table*}

The system successfully designed effective strategies for all selected categories from Nemotron-CC\citep{karimi2025nemotroncc} spanning 672B tokens across academic content in mathematics, computer science, medicine, and other STEM fields, each at two quality levels. Applying the optimized strategies produces $\text{Nemotron-CC}_{\text{ASI+}}$ (504B tokens). Training 3B-parameter models from scratch on 500B tokens and evaluating across 18 benchmarks, $\text{Nemotron-CC}_{\text{ASI+}}$ achieves 44.13 average score, surpassing raw data by 3.96 points and established corpora including DCLM, FineWeb-Edu, and Ultra-FineWeb under identical training budgets. Gains are particularly pronounced on knowledge-intensive tasks: \textbf{MMLU +18.64 points, CSQA +18.80 points, MedQA +13.48 points}.

\paragraph{Analysis}
We analyze the design characteristics of discovered strategies. Across all categories, the system converges on cleaning-focused approaches without any prescriptive guidance on which operations to employ, consistently combining targeted noise removal (HTML artifacts, duplicates, PII), format normalization (whitespace, punctuation), and domain-aware preservation rules. This convergence demonstrates that systematic cleaning with domain-specific preservation suffices for substantial quality improvements. Beyond this shared foundation, effective strategies exhibit consistent design patterns: concrete criteria with measurable thresholds, targeted deletion of specific elements, and explicit preservation rules that prevent over-aggressive filtering. The 2.93-point gap between optimized and suboptimal strategies further illustrates the value of iterative refinement. Rather than relying on one-shot generation, the agentic evolution process accumulates diagnostic feedback across iterations, analyzing what worked, what gaps remain, and how to improve, enabling the system to navigate the vast strategy space and converge on high quality solutions.

%% file: exp_rl.tex
\subsection{Scenario 3: Reinforcement Learning Algorithm Design}

\paragraph{Task Formulation} In this phase, we tasked the Evolve system with designing a novel Reinforcement Learning (RL) algorithm for Large Language Model (LLM) training. Using Group Relative Policy Optimization (GRPO) as the baseline, the objective was to redesign the mechanism for advantage allocation across sequences and the subsequent gradient computation. To succeed, the system was required to comprehend the mathematical foundations of RL, interpret diverse training metrics, and distinguish between stochastic training instability and genuine algorithmic improvements.

\paragraph{Methodology}

We initialized the cognition repository with 10 high-quality papers published subsequent to GRPO, covering variance reduction techniques and KL-penalty modifications. These entries provided the system with a preliminary understanding of the current research frontier, constraining the search space toward plausible mathematical directions while avoiding theoretical dead ends. We employed a two-stage validation protocol to balance computational cost and evaluation reliability. In the exploration phase, candidate algorithms were trained on a 4B parameter model for 150 steps and evaluated on 6 mathematics benchmarks. Promising candidates were then scaled to a 14B parameter model for 300 steps in the verification phase, with the evaluation suite expanded to Abstract Reasoning, STEM, Finance, and Coding domains to test generalization. For fitness scoring, we simplified the function relative to the architecture search by removing sigmoid normalization, instead using a linear weighted sum of benchmark accuracy and LLM-as-a-Judge qualitative scores, focusing directly on raw performance gains and algorithmic coherence.

\paragraph{Results}

Over the course of 300 evolutionary rounds, the system trained and evaluated a diverse array of policy gradient modifications, yielding 10 algorithms that outperformed the GRPO baseline in the exploration phase. Upon scaling to the 14B parameter verification phase, 3 algorithms demonstrated statistically significant improvements across all tested domains. On mathematical benchmarks, the best evolved variants improve over GRPO by +12.5 points on AMC32 (67.5 $\rightarrow$ 80.0), +11.67 points on AIME24 (20.00 $\rightarrow$ 31.67), and +5.04 points on OlympiadBench (45.92 $\rightarrow$ 50.96), suggesting that the system can effectively optimize subtle mathematical trade-offs in loss function design.

\begin{table}[t]
    \centering
    \small
    \setlength{\tabcolsep}{6pt}
    \begin{tabular}{lccccc}
    \toprule
    \textbf{Method} & \textbf{Math500} & \textbf{AMC32} & \textbf{AIME25} & \textbf{AIME24} & \textbf{OlympiadBench} \\
    \midrule
    GRPO$_{\text{Human}}$                             & 82.0 & 67.5 & 20.00 & 20.00 & 45.92 \\
    \midrule
    Algorithm1$_{\text{ASI-Evolve}}$                       & 85.6 & 80.0 & 29.58 & 31.67 & 50.96 \\
    Algorithm2$_{\text{ASI-Evolve}}$                       & 84.6 & 77.5 & 23.75 & 23.33 & 48.74 \\
    Algorithm3$_{\text{ASI-Evolve}}$                       & 84.8 & 77.5 & 30.00 & 30.00 & 49.18 \\
    Algorithm4$_{\text{ASI-Evolve}}$                       & 82.4 & 75.0 & 20.00 & 20.00 & 46.81 \\
    Algorithm5$_{\text{ASI-Evolve}}$                       & 82.0 & 72.5 & 23.33 & 20.00 & 45.62 \\
    \bottomrule
    \end{tabular}
    \caption{Performance comparison of evolved RL algorithms and the GRPO baseline on mathematical reasoning benchmarks. All methods were trained on Qwen-3-14B-base within the SIIRL framework using the Skywork OR1 dataset and evaluated after 250 training steps.}
    \label{tab:rl_algorithm_math_results}
    \end{table}

\paragraph{Analysis}
We highlight two representative high-performing algorithms that exhibit distinct theoretical innovations. (1) \textit{Algorithm A (Pairwise Asymmetric Optimization)} introduces a \textit{comparative advantage estimation}: instead of using a group mean, the advantage for a response $A$ is calculated by averaging the $\tanh$-normalized pairwise reward differences against all other group samples ($R_A - R_B$). It further employs an \textit{asymmetric clipping mechanism} that dynamically adjusts the PPO clipping window $[\epsilon_{down}, \epsilon_{up}]$ based on the sign of the advantage, and implements \textit{High-Impact Gradient Dropout}, stochastically masking gradients for the most influential tokens (highest probability $\times$ advantage) to prevent overfitting to specific keywords. (2) \textit{Algorithm B (Budget-Constrained Dynamic Radius)} adopts percentile-based normalization for advantage calculation, defined as $(r - c)/s$. Its core innovation is the \textit{Global Update Budget} ($z_{cap}$): the algorithm dynamically assigns each token a trusted update radius inversely proportional to the magnitude of its advantage, and strictly enforces the exponential bound $\exp(c) \times |A| \leq z_{cap}$, mathematically guaranteeing that the total policy update magnitude remains within a pre-defined budget and effectively stabilizing training on noisy data. These two algorithms illustrate that ASI-EVOLVE can perform rigorous mathematical derivation and discover principled solutions to fundamental stability and variance challenges in RL training, paralleling innovations seen in human-designed algorithmic advances.

%% file: ablation.tex
\section{Empirical Analysis}
\label{sec:ablation}

Having demonstrated ASI-Evolve's effectiveness across four complex real-world domains in Section~\ref{sec:main-tasks}, we now turn to a systematic empirical analysis of the framework. We first evaluate ASI-Evolve's performance on the circle packing task---a task that has been adopted by multiple evolutionary frameworks and thus enables direct comparison---to assess how our system compares against existing approaches and to examine the impact of design choices such as base model and sampling algorithm. We then conduct controlled ablation studies on the same task to isolate the contribution of individual components. Finally, we further demonstrate that the solutions produced by ASI-Evolve can be genuinely applied beyond the AI/ML stack: the model architecture evolved by ASI-Evolve achieves strong results in a biomedical setting, showing that AI-optimized designs carry practical value in real-world domains.

\subsection{Benchmarking ASI-Evolve on Circle Packing}
\label{subsec:circle_packing}

\paragraph{Task.} We use the circle packing task from AlphaEvolve as a controlled evaluation platform. The problem requires placing 26 circles within a $1\times1$ square to maximize the sum of their radii. This is a classic combinatorial optimization problem with low verification cost, yet it still demands non-trivial algorithm design and iterative refinement, making it a suitable proxy for comparing evolutionary frameworks under aligned conditions.

\paragraph{Key results at a glance.} Table~\ref{tab:circle_packing} summarizes results across representative evolutionary frameworks. Our ASI-Evolve reaches 2.63597 in as few as 17 steps---the fastest among all compared systems---and achieves a best score of 2.635983, comparable to the top results reported by other frameworks.

\begin{table}[h]
\centering
\setlength{\tabcolsep}{8pt}
\begin{tabular}{llcc}
\toprule
\textbf{Framework} & \textbf{Model} & \textbf{Rounds} & \textbf{Best Score} \\
\midrule
AlphaEvolve~\citep{novikov2025alphaevolvecodingagentscientific} & Gemini 2.0 Flash + Claude 3.7 & --- & 2.6359 \\
OpenEvolve~\citep{openevolve}   & Gemini 2.0 Flash + Claude 3.7 & 460 & 2.6343 \\
LoongFlow~\citep{loongflow2024} & DeepSeek-R1-250528             & --- & 2.6360 \\
SkyDiscover                     & GPT-5                          & 89  & 2.6360 \\
\midrule
ASI-Evolve \textbf{(Ours)}      & GPT-5-mini                     & \textbf{17}  & \textbf{2.6360} \\
\bottomrule
\end{tabular}
\caption{Circle packing comparison across evolutionary frameworks (26 circles in a $1\times1$ unit square; higher sum of radii is better). ``Rounds'' denotes the number of evolution steps to reach the reported best score; ``---'' indicates the value was not reported.}
\label{tab:circle_packing}
\end{table}

\subsection{Comparison Experiments}
\label{subsec:ablation_comparison}

\subsubsection{Framework Comparison}

Using Qwen3-32B as the base model, we compare ASI-Evolve against two representative evolutionary frameworks---OpenEvolve and GEPA---under an aligned prompt setup. As shown in Figure~\ref{fig:ablation_comparison}(a), the three frameworks exhibit noticeably different evolution dynamics:

\begin{itemize}[noitemsep,topsep=2pt]
\item \textbf{OpenEvolve} continues to evolve throughout the run, but exhibits high variance across independent runs and delivers only limited overall improvement, with scores plateauing well below the SOTA level.
\item \textbf{GEPA} achieves competitive scores, converging to a range around 2.630. Its performance is substantially better than OpenEvolve, reflecting the benefit of structured evolutionary design.
\item \textbf{ASI-Evolve} exits the cold-start phase with a noticeably higher score than both baselines, continues to improve steadily throughout the run, and is the only framework to reliably reach SOTA-level performance.
\end{itemize}

These findings align with our ablation results: the Analyzer and Cognition components provide structured feedback and domain priors that translate directly into faster convergence and higher final scores, advantages that persist across diverse evolutionary frameworks.

\subsubsection{Model Comparison}

\begin{figure}[t]
  \centering
  \begin{minipage}[t]{0.48\linewidth}
    \centering
    \includegraphics[width=\linewidth]{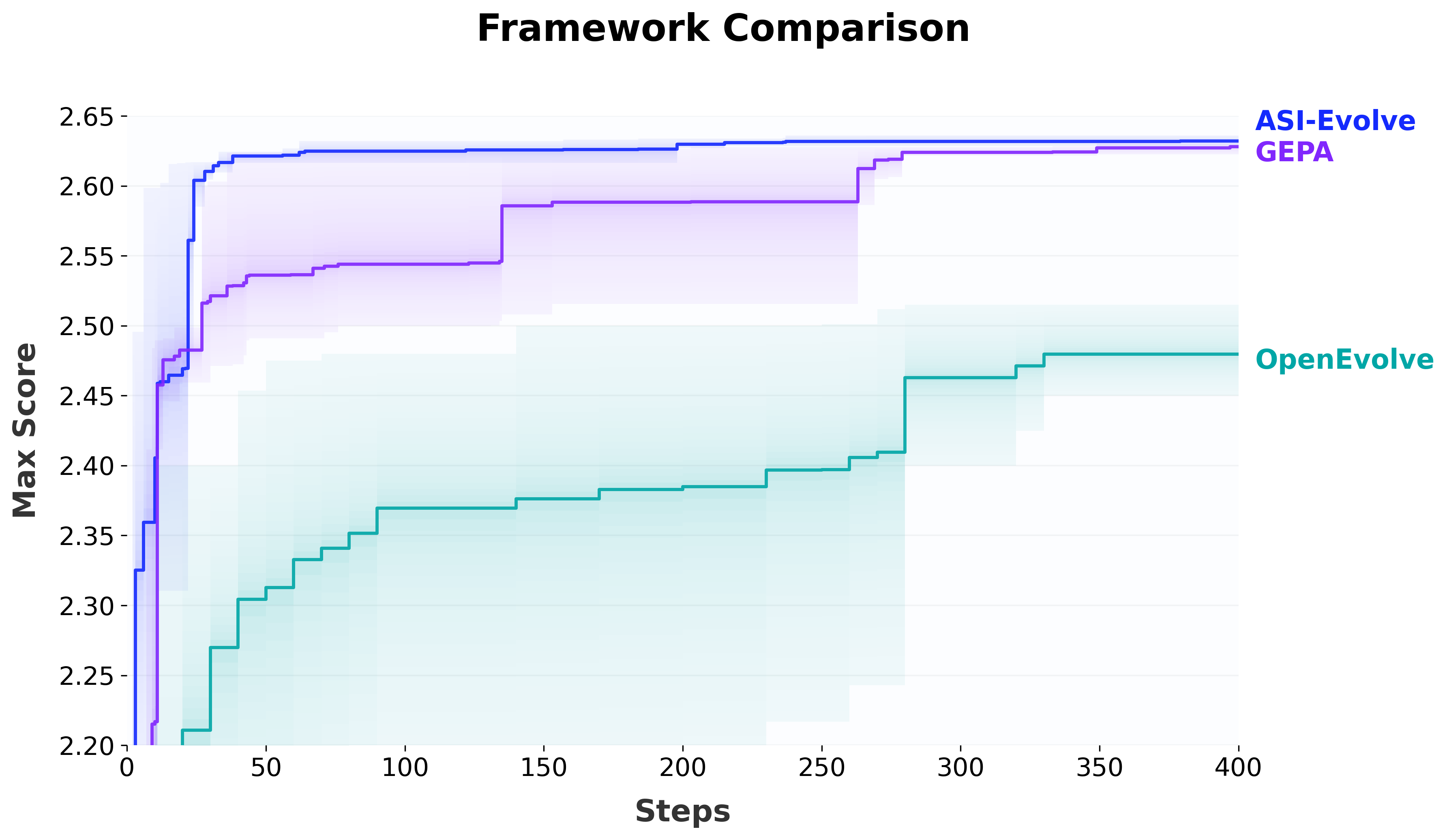}
    \caption*{\textbf{(a)} ASI-Evolve vs.\ GEPA vs.\ OpenEvolve.}
  \end{minipage}\hfill
  \begin{minipage}[t]{0.48\linewidth}
    \centering
    \includegraphics[width=\linewidth]{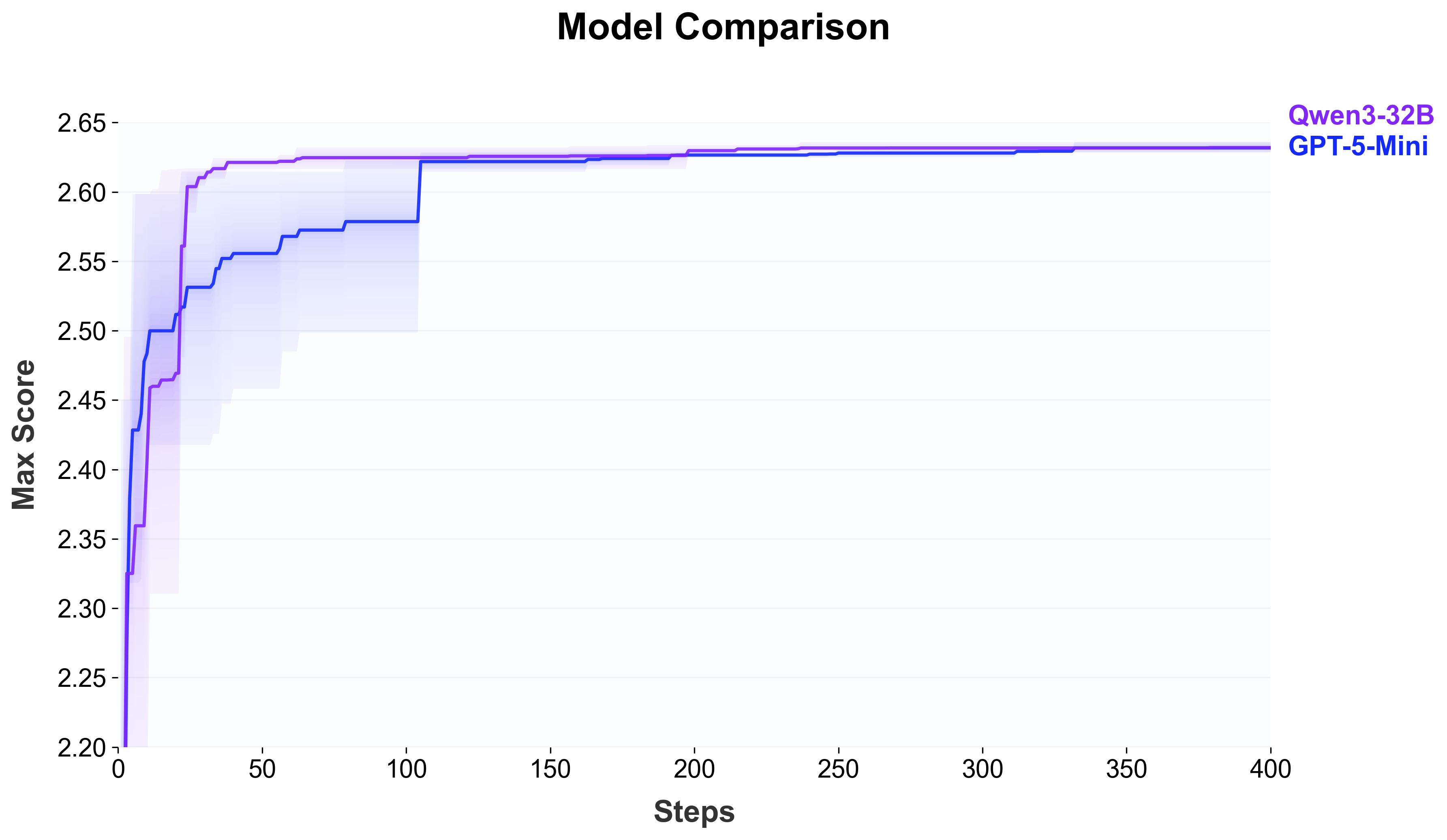}
    \caption*{\textbf{(b)} Base-model comparison.}
  \end{minipage}
  \caption{\textbf{Comparison experiments.} \textbf{(a)} Evolution curves for ASI-Evolve, GEPA, and OpenEvolve. \textbf{(b)} Evolution curves for ASI-Evolve using GPT-5-mini and Qwen3-32B. Shaded regions indicate variability across repeated runs.}
  \label{fig:ablation_comparison}
\end{figure}

We further run ASI-Evolve with GPT-5-mini and Qwen3-32B as the base model. As shown in Figure~\ref{fig:ablation_comparison}(b), the two runs converge to a similar range and exhibit highly consistent mid-to-late-stage improvement trends, indicating that the framework's evolution capability is not tied to a particular model family. At the same time, the early-stage cadence can differ across models: one model may enter the high-score regime earlier, while the other catches up via a distinct jump in the mid stage and ultimately converges to a comparable level. This further supports ASI-Evolve's compatibility and robustness across base models.

\subsubsection{Algorithm Comparison}

The database sampling algorithm determines how parent nodes are selected each round, directly shaping the balance between exploration and exploitation. \textbf{MAP-Elites} maintains a quality-diversity archive partitioned by behavioral features, actively preserving diverse niches to prevent premature convergence and encourage broad coverage of the solution space. \textbf{UCB1} treats each node as a bandit arm and selects based on an upper confidence bound that combines estimated value with an exploration bonus inversely proportional to visit count---rewarding high-scoring nodes while still visiting under-explored ones. \textbf{Random} sampling selects parent nodes uniformly at random from the database, without any preference for score or diversity.

\begin{wrapfigure}[17]{r}{0.48\linewidth}
  \vspace{-0.5em}
  \centering
  \includegraphics[width=\linewidth]{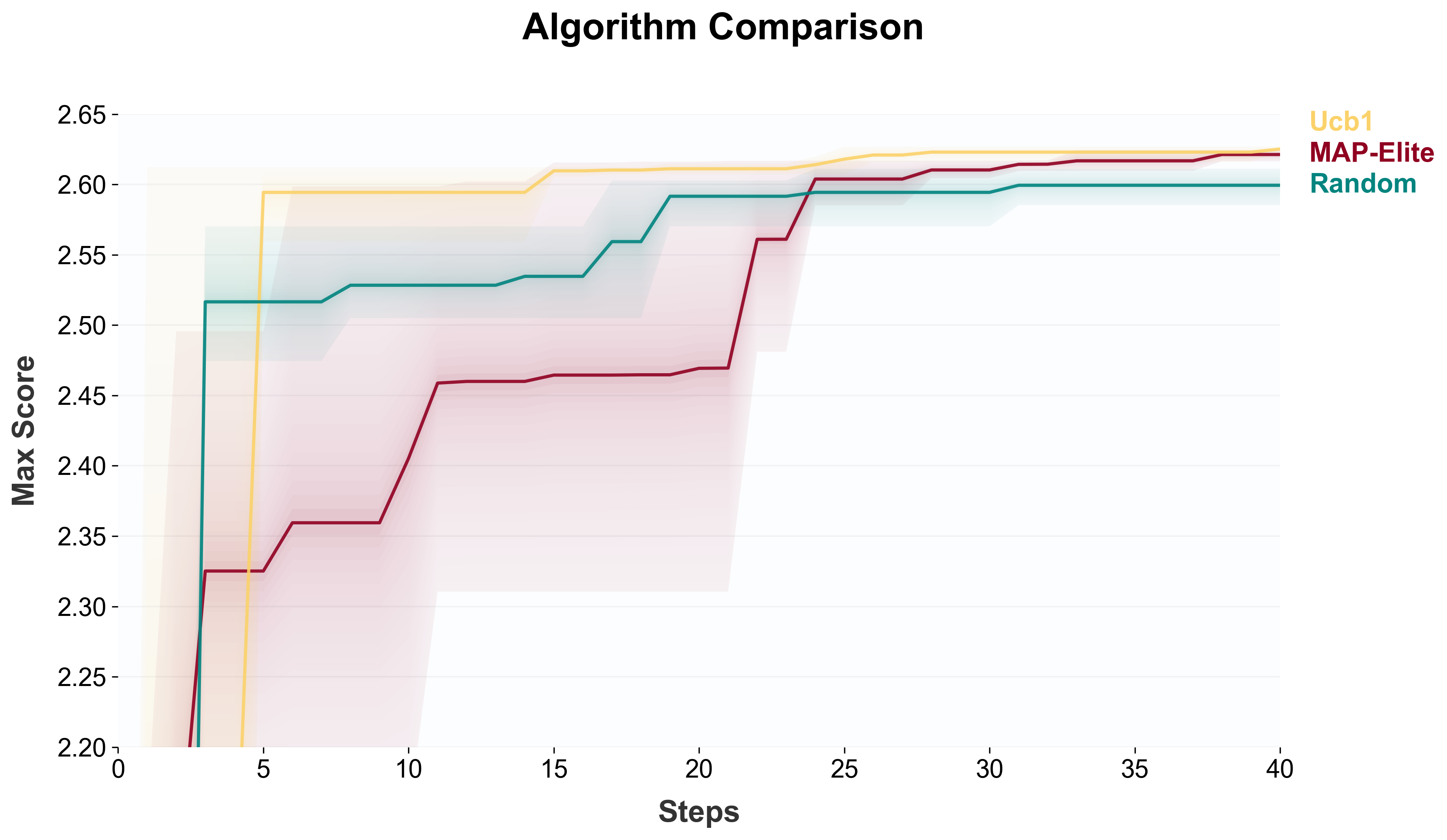}
  \caption{\textbf{Sampling algorithm comparison.} Evolution curves for ASI-Evolve with MAP-Elites, UCB1, and Random sampling on Qwen3-32B. Shaded regions show run-to-run variability.}
  \label{fig:comparison_algorithm}
\end{wrapfigure}

As shown in Figure~\ref{fig:comparison_algorithm}, the three sampling strategies exhibit distinct dynamics. \textbf{Random} achieves a higher initial score than MAP-Elites in early steps---because it places no diversity constraint on parent selection, the Researcher is free to immediately exploit the most informative nodes without being redirected toward under-explored regions. However, this early advantage erodes over time: without any mechanism to balance exploration and exploitation, Random sampling gradually slows its rate of improvement and falls behind both MAP-Elites and UCB1 in later stages.

\textbf{UCB1} reaches high-score regions faster than MAP-Elites and exhibits lower variance across runs. This may appear counterintuitive---UCB1's exploitation bias could, in principle, cause it to converge prematurely on a narrow set of high-scoring parents and miss the broader diversity that MAP-Elites is designed to preserve. We attribute the reversed outcome to the role of cognition: with a well-initialized cognition repository and structured Analyzer feedback already providing directional guidance, the additional diversity enforced by MAP-Elites becomes less valuable, while UCB1's value-guided selection allows the system to rapidly concentrate on productive design patterns. These results indicate that, in the presence of cognition, the system can relax its dependence on diversity-preserving samplers and thereby achieve faster convergence. Notably, combining UCB1 with GPT-5-mini, the system discovered a circle-packing solution scoring 2.63597---matching the SOTA level---in just 17 steps. By contrast, MAP-Elites with the same GPT-5-mini base model required 79 steps to reach an equivalent score (2.63597), illustrating how exploitation-oriented sampling, when guided by strong cognition priors, can dramatically accelerate the discovery of high-quality solutions.

\subsection{Ablation Study: Validating Component Effectiveness}
\label{subsec:ablation_self}

\subsubsection{Ablation design}

We design the following controlled experiments to systematically evaluate key components of the ASI-Evolve framework:

\begin{enumerate}[noitemsep,topsep=2pt]
\item \textbf{Full Method:} \modelname with Analyzer, Cognition repository, and the complete four-stage loop (Learn--Design--Experiment--Analyze).
\item \textbf{No Analyzer:} Remove the Analyzer module. After the Engineer runs experiments, raw evaluation scores and execution logs are stored directly in the Database as results for the next Researcher iteration.
\item \textbf{No Cognition:} Remove the Cognition repository. The Researcher receives no literature-derived prior knowledge; the system relies entirely on self-driven trial-and-error learning.
\end{enumerate}

Considering the high variance inherent to evolutionary systems, we run each configuration three times independently, and analyze the overall improvement rate, convergence behavior, and cross-run gaps from evolution curves.

\begin{figure}[t]
  \centering
  \includegraphics[width=0.85\linewidth]{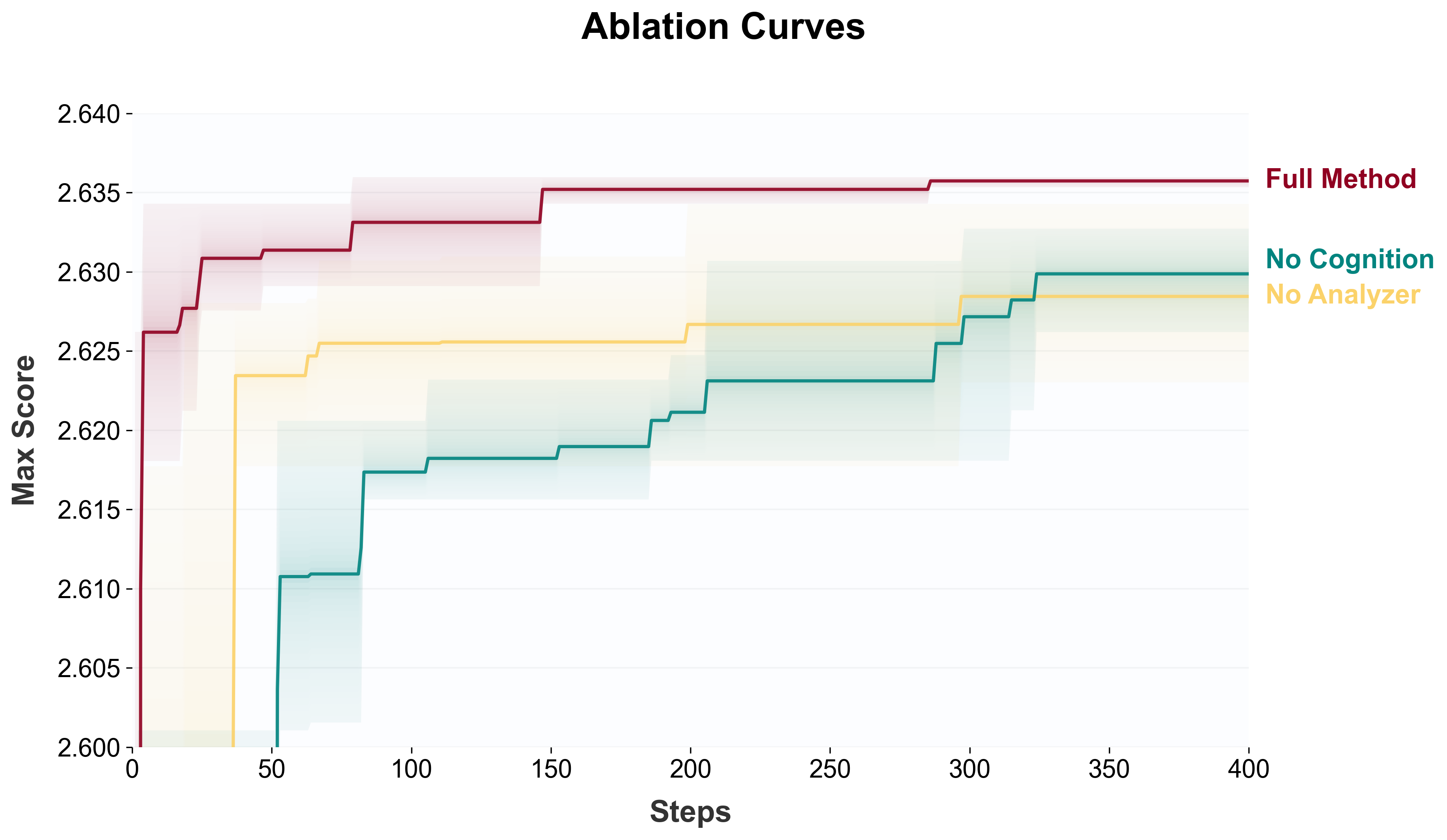}
  \caption{\textbf{Ablation on circle packing.} Evolution curves for the full method and ablated variants. Shaded regions indicate variability across repeated runs.}
  \label{fig:ablation_curves}
\end{figure}

\subsubsection{Impact of removing Analyzer}

Figure~\ref{fig:ablation_curves} compares evolution curves with and without the Analyzer. We observe:

\paragraph{High initial scores attributable to Cognition.} Even without the Analyzer, the No Analyzer variant begins from a relatively high score in the early phase. We attribute this to the Cognition repository: domain priors from literature guide the Researcher toward promising directions from the outset, regardless of whether structured feedback is available. This confirms that Cognition provides a meaningful cold-start advantage independently of Analyzer-driven feedback.

\paragraph{Long plateau with limited sustained improvement.} Despite the high starting point, the No Analyzer variant subsequently enters a prolonged plateau where further iterations yield only marginal gains. The ability to continuously push toward higher ceilings becomes markedly weaker, and improvements become sporadic and less reproducible. Notably, some runs do eventually reach SOTA-level scores; we attribute this to two factors: first, even without a dedicated Analyzer, the system still receives a limited feedback signal through raw evaluation scores, providing some directional guidance; second, the circle packing task is relatively straightforward, so the performance gap between configurations is less pronounced than it would be on more complex tasks. Nevertheless, the absence of structured analysis consistently leads to slower and less reliable sustained evolution compared to the full method.

\subsubsection{Impact of removing Cognition}

Figure~\ref{fig:ablation_curves} also includes evolution curves with and without the Cognition repository. Key observations:

\paragraph{Delayed exploration onset.} The No Cognition variant exhibits a more pronounced cold-start cost: early improvements are slower and less stable, and the curve can remain in a relatively low-score region for a prolonged period. After sufficient effective experience is accumulated, the curve shows a noticeable jump and then gradually enters a higher-scoring, productive exploration regime. This matches the intended role of Cognition: it does not change the framework's core learning mechanism, but provides better priors to reduce unproductive exploration and shorten the ``trial'' phase.

\paragraph{Sustained evolution capability.} Despite a slower start, the No Cognition variant still maintains effective evolution capability. Even without external priors, ASI-Evolve's core mechanisms can continuously learn from self-guided trial-and-error and gradually distill effective strategies. This also implies that in entirely novel domains where effective priors are unavailable, the framework remains usable, albeit requiring longer exploration.

%% file: exp_medical.tex
% exp_medical.tex - Biomedical Target Discovery Experiment
% This file contains the Biomedical Target Discovery experiment section

\subsection{Validating Real-World Applicability: Drug–Target Interaction Discovery}
\label{subsec:medical}

The experiments above confirm that ASI-Evolve delivers strong results on AI-for-AI tasks and that its components contribute meaningfully to performance. Yet a legitimate concern persists: even if AI can effectively optimize AI systems, whether the resulting solutions are genuinely useful when deployed in the real world. To address this directly, we present results on drug-target interaction (DTI) prediction, where the architecture evolved by ASI-Evolve is applied to a biomedical task to demonstrate that AI-optimized designs carry practical value beyond the AI/ML domain.

\paragraph{Task Formulation}
We apply ASI-Evolve to Drug-Target Interaction (DTI) prediction, a central problem in AI-driven drug discovery. Effective DTI models must simultaneously capture modality-specific representations of drug molecules and protein targets, as well as their complex interaction patterns. The architectural design space is large and discrete---spanning feature extraction, fusion mechanisms, and interaction modeling---with limited theoretical guidance. We use DrugBAN~\cite{DrugBAN} as the seed architecture and aim to discover improved variants through automated architectural evolution.

\paragraph{Methodology}
The cognition repository is initialized from approximately 80 papers on graph neural networks, attention mechanisms, and DTI modeling, capturing known limitations such as over-reliance on shallow cross-attention and insufficient higher-order interaction modeling. In each round, the Researcher proposes and implements candidate architectural modifications (e.g., restructuring the drug-protein interaction module or introducing new cross-modal fusion strategies); the Engineer trains the modified DrugBAN on the BindingDB development set; and the Analyzer evaluates performance across benchmark splits and metrics (AUROC, AUPRC, F1, MCC), flagging failure modes such as sensitivity to protein length or chemical scaffold diversity. We evaluate on four datasets (BindingDB, Human, BioSNAP, C.elegans) under four generalization settings: \textbf{random split}, \textbf{unseen drug}, \textbf{unseen protein}, and \textbf{unseen drug and protein}. Fitness combines AUROC and AUPRC as primary objectives with F1 and MCC as secondary criteria. Over 100+ evolution rounds, 100+ candidate architectures were evaluated.

\paragraph{Results.} Table~\ref{tab:dti_results} summarizes the performance of our best discovered architecture compared to DrugBAN baseline and six state-of-the-art baselines across multiple benchmark datasets and evaluation settings.

\begin{table}[t]
\centering \small
\resizebox{\columnwidth}{!}{%
\begin{tabular}{@{}llcccccccc@{}}
\toprule
& & \multicolumn{2}{c}{\textbf{BindingDB-Dev}} & \multicolumn{2}{c}{\textbf{BindingDB-Random}} & \multicolumn{2}{c}{\textbf{Human}} & \multicolumn{2}{c}{\textbf{BioSNAP}} \\
\textbf{Discovered by} & \textbf{Model} & AUROC & F1 & AUROC & F1 & AUROC & F1 & AUROC & F1 \\
\midrule
\multirow{5}{*}{Human \faUserGraduate} & TransformerCPI & -     & -     & 93.96 & 87.02 & 96.03 & 91.35 & 86.35 & 78.53 \\
                       & PSICHIC        & -     & -     & 91.67 & 83.30 & 98.55 & 94.85 & 91.64 & 84.87 \\
                       & ConPlex        & -     & -     & 93.59 & 83.39 & 97.04 & 90.38 & 88.40 & 72.86 \\
                       & ColdStartCPI   & -     & -     & 89.27 & 85.27 & 98.29 & 93.57 & 93.39 & 86.84 \\
\cmidrule{2-10}
                       & DrugBAN        & 94.15 & 86.89 & 94.89 & 87.96 & 98.61 & 95.40 & 89.43 & 82.69 \\
\midrule
AI  \faRobot                   & \textbf{ASI-Evolve}  & \textbf{96.06} & \textbf{89.84} & \textbf{95.94} & \textbf{89.35} & \textbf{98.89} & \textbf{95.32} & \textbf{89.68} & \textbf{82.92} \\
\bottomrule
\end{tabular}%
}
\caption{Performance comparison on drug-target interaction prediction across multiple benchmarks. We report AUROC (\%) and F1 (\%) scores for key evaluation settings. Bold indicates best performance.}
\label{tab:dti_results}
\end{table}

% \begin{table}[t]
% \centering \small
% \resizebox{\columnwidth}{!}{%
% \begin{tabular}{@{}l|cc|cc|cc|cc@{}}
% \toprule
% & \multicolumn{2}{c|}{\textbf{BindingDB-Dev}} & \multicolumn{2}{c|}{\textbf{BindingDB-Random}} & \multicolumn{2}{c|}{\textbf{Human}} & \multicolumn{2}{c}{\textbf{BioSNAP}} \\
% \textbf{Model} & AUROC & F1 & AUROC & F1 & AUROC & F1 & AUROC & F1 \\
% \midrule
% TransformerCPI & - & - & 0.9396 & 0.8702 & 0.9603 & 0.9135 & 0.8635 & 0.7853 \\
% PSICHIC & - & - & 0.9167 & 0.8330 & 0.9855 & 0.9485 & 0.9164 & 0.8487 \\
% ConPlex & - & - & 0.9359 & 0.8339 & 0.9704 & 0.9038 & 0.8840 & 0.7286 \\
% ColdStartCPI & - & - & 0.8927 & 0.8527 & 0.9829 & 0.9357 & 0.9339 & 0.8684 \\
% \midrule
% DrugBAN (baseline) & 0.9415 & 0.8689 & 0.9489 & 0.8796 & 0.9861 & 0.9540 & 0.8943 & 0.8269 \\
% \textbf{Ours} & \textbf{0.9606} & \textbf{0.8984} & \textbf{0.9594} & \textbf{0.8935} & \textbf{0.9889} & \textbf{0.9532} & \textbf{0.8968} & \textbf{0.8292} \\
% \midrule
% \textit{Improvement} & \textit{+0.0191} & \textit{+0.0295} & \textit{+0.0105} & \textit{+0.0139} & \textit{+0.0028} & \textit{-0.0008} & \textit{+0.0025} & \textit{+0.0023} \\
% \bottomrule
% \end{tabular}%
% }
% \caption{Performance comparison on drug-target interaction prediction across multiple benchmarks. We report AUROC and F1 scores for key evaluation settings. Bold indicates best performance.}
% \label{tab:dti_results}
% \end{table}

Our discovered architecture achieves consistent improvements over the DrugBAN baseline across most evaluation settings. On the BindingDB development set, we observe a substantial improvement of +1.91 AUROC points (0.9415 → 0.9606) and +2.95 F1 points (0.8689 → 0.8984), demonstrating effective in-distribution learning. Importantly, these improvements transfer to the test splits: on BindingDB-Random, our method achieves +1.05 AUROC points and +1.39 F1 points improvement. On the challenging Human benchmark, our architecture maintains strong performance with 0.9889 AUROC, while on BioSNAP it achieves modest but consistent gains (+0.25 AUROC, +0.23 F1).

\textit{Generalization Analysis.} We further analyze performance on cold-start scenarios where the model must generalize to completely unseen drugs or proteins (Table~\ref{tab:dti_coldstart}). These settings are particularly challenging as they require the model to extract transferable patterns from molecular structures rather than memorizing specific drug-target pairs.

\begin{table}[t]
\centering
\tiny
\resizebox{\columnwidth}{!}{%
\begin{tabular}{@{}llcccccc@{}}
\toprule
& & \multicolumn{2}{c}{\textbf{Unseen Drug}} & \multicolumn{2}{c}{\textbf{Unseen Protein}} & \multicolumn{2}{c}{\textbf{Unseen Drug and Protein}} \\
\textbf{Discovered by} & \textbf{Model} & AUROC & F1 & AUROC & F1 & AUROC & F1 \\
\midrule
Human \faUserGraduate & DrugBAN       & 79.15 & 77.39 & 82.26 & 75.76 & 76.47 & 71.53 \\
\midrule
AI \faRobot   & \textbf{Ours} & \textbf{86.09} & \textbf{82.35} & \textbf{85.82} & \textbf{78.44} & \textbf{80.83} & \textbf{74.51} \\
\bottomrule
\end{tabular}%
}
\caption{Cold-start performance comparison. We report AUROC (\%) and F1 (\%) scores. Models must predict interactions for unseen drugs, proteins, or both.}
\label{tab:dti_coldstart}
\end{table}

% \begin{table}[t]
% \centering 
% \tiny
% \resizebox{\columnwidth}{!}{%
% \begin{tabular}{@{}l|cc|cc|cc@{}}
% \toprule
% & \multicolumn{2}{c|}{\textbf{Unseen Drug}} & \multicolumn{2}{c|}{\textbf{Unseen Protein}} & \multicolumn{2}{c}{\textbf{Unseen Drug and Protein}} \\
% \textbf{Model} & AUROC & F1 & AUROC & F1 & AUROC & F1 \\
% \midrule
% DrugBAN (baseline) & 0.7915 & 0.7739 & 0.8226 & 0.7576 & 0.7647 & 0.7153 \\
% \textbf{Ours} & \textbf{0.8609} & \textbf{0.8235} & \textbf{0.8582} & \textbf{0.7844} & \textbf{0.8083} & \textbf{0.7451} \\
% \midrule
% \textit{Improvement} & \textit{+0.0694} & \textit{+0.0496} & \textit{+0.0356} & \textit{+0.0268} & \textit{+0.0436} & \textit{+0.0298} \\
% \bottomrule
% \end{tabular}%
% }
% \caption{Cold-start performance comparison. Models must predict interactions for unseen drugs, proteins, or both.}
% \label{tab:dti_coldstart}
% \end{table}

The results reveal substantial generalization improvements: +6.94 AUROC points for unseen drugs, +3.56 points for unseen proteins, and +4.36 points in the doubly-cold-start setting. These improvements significantly exceed the in-distribution gains, suggesting that the evolved architecture has learned more robust and transferable representations of molecular interactions.

\paragraph{Analysis.}
The best discovered architecture (ban\_sinkhorn\_ds\_marginal\_topk\_v6) introduces three key innovations over DrugBAN. (1)~\textbf{Sinkhorn Attention}: replacing standard bilinear attention with optimal-transport-based Sinkhorn iterations enforces doubly-stochastic constraints, ensuring balanced attention allocation between drug and protein features and preventing attention collapse. (2)~\textbf{Domain-Specific Marginalization}: specialized marginalization over molecular substructures (drugs) and protein domains aggregates interaction patterns across distinct semantic spaces, enabling more compositional modeling of binding mechanisms. (3)~\textbf{Top-k Sparse Gating}: learnable top-k selection dynamically focuses on the most relevant interaction patterns, reducing noise from irrelevant molecular features. These choices align with domain knowledge—optimal transport has theoretical connections to binding affinity~\cite{ot_drug}, while compositional reasoning over substructures reflects established principles in medicinal chemistry.

Tracking the evolution process reveals that early iterations draw heavily on cognition from graph attention and molecular representation papers; as experiments accumulate, the system synthesizes cross-paper insights—most notably, the Sinkhorn mechanism emerged from combining optimal transport theory with bipartite matching concepts from computational biology. Consistent with this pattern, the fitness curve improves steadily overall, with notable jumps often appearing when cross-domain ideas are integrated successfully. Regarding common bottlenecks during evolution, the Analyzer repeatedly identifies failure modes such as representational collapse, overfitting to binding sites, and attention saturation, and the system progressively mitigates them through entropy regularization and orthogonality constraints.

Compared to expert-designed methods (TransformerCPI~\cite{transformercpi}, PSICHIC~\cite{psichic}, ConPlex~\cite{conplex}, ColdStartCPI~\cite{coldstartcpi}) that rely on pre-trained molecular encoders, protein language models, or specialized graph convolutions, our autonomously discovered architecture achieves competitive or superior performance across benchmarks. These results indicate that AI-driven architectural evolution can deliver real gains on challenging cross-domain tasks.

%% file: appendix/exp_ablation.tex
\section{Analysis Configuration Details}
\label{app:ablation_config}

This appendix summarizes the concrete experiment settings used in the circle-packing analysis section.

\paragraph{Framework and model comparison (Section~\ref{subsec:ablation_comparison}).}
\begin{itemize}[noitemsep,topsep=2pt]
\item \textbf{Compared settings:} aligned comparison with OpenEvolve-style prompting, and backbone comparison between GPT-5-mini and Qwen3-32B under the same system design.
\item \textbf{Alignment protocol:} prompt style, full-file evolution mode, MAP-Elites/island database, and population-related settings were aligned to the OpenEvolve circle-packing setup.
\item \textbf{GPT-5-mini setting:} temperature $0.7$, top-$p=0.95$, max tokens $32768$.
\item \textbf{Qwen3-32B setting:} temperature $0.6$, top-$p=0.95$, max tokens $65536$, seed $=42$, thinking enabled, top-$k=20$, and min-$p=0$.
\item \textbf{Shared system setting:} max size $=70$, \texttt{sample\_n=3}, engineer timeout $=300$s, 4 parallel workers, cognition retrieval top-$k=5$, Judge disabled, and 3 runs per model.
\end{itemize}

\paragraph{Sampling algorithm comparison (Section~\ref{subsec:ablation_comparison}).}
\begin{itemize}[noitemsep,topsep=2pt]
\item \textbf{Backbone model:} Qwen3-32B with the same decoding setting as above: temperature $0.6$, top-$p=0.95$, max tokens $65536$, seed $=42$, thinking enabled, top-$k=20$, and min-$p=0$.
\item \textbf{Controlled factors:} prompt template, cognition setting, full-file evolution mode, engineer timeout, parallelism, and database size were held fixed.
\item \textbf{Only changed factor:} database sampling algorithm. MAP-Elites uses the island configuration above, while UCB1 uses \texttt{algorithm=ucb1} with exploration constant $c=1.414$.
\item \textbf{Repeats:} 3 independent runs per sampling strategy.
\end{itemize}

\paragraph{Ablation study (Section~\ref{subsec:ablation_self}).}
\begin{itemize}[noitemsep,topsep=2pt]
\item \textbf{Base model:} GPT-5-mini with temperature $0.7$, top-$p=0.95$, and max tokens $32768$.
\item \textbf{System loop:} Researcher--Engineer--Analyzer enabled by default; the ``No Analyzer'' setting disables Analyzer only, while ``No Cognition'' removes cognition usage but keeps the rest of the loop unchanged.
\item \textbf{Researcher setting:} full-file generation, max code length $100000$, and up to 3 retries.
\item \textbf{Engineer setting:} timeout $=300$s, up to 2 retries, and 4 parallel workers.
\item \textbf{Database setting:} max size $=70$, sampling algorithm = island-based MAP-Elites, 5 islands, migration interval $=10$, migration rate $=0.1$, exploration ratio $=0.2$, exploitation ratio $=0.6$, feature dimensions = \{\texttt{complexity}, \texttt{diversity}\}, and 10 bins per feature.
\item \textbf{Cognition setting:} retrieval top-$k=5$, score threshold $=0.4$, sentence-transformer embedding dimension $=384$, and web search disabled.
\item \textbf{Other shared settings:} \texttt{sample\_n=3}, Judge disabled, and 3 independent runs per condition. All curves report the mean trend with variability across runs.
\end{itemize}

\paragraph{Cognition repository contents for circle packing.}
\label{app:cognition_contents}
All experiments that use cognition share a common knowledge base initialized before the first evolution step. The 12 cognition items span three categories:

\begin{itemize}[noitemsep,topsep=2pt]
\item \textbf{Geometric priors (4 items).}
  \begin{itemize}[noitemsep,topsep=1pt]
  \item \emph{Hexagonal close packing:} theoretical density $\pi/(2\sqrt{3}) \approx 0.9069$ for infinite planes; unit-square boundary effects reduce achievable density; best patterns use hexagonal arrangements with careful corner and boundary handling.
  \item \emph{Edge and corner effects:} circles near boundaries are space-constrained; larger circles should be placed at corners and edges (corner radius up to $\frac{\sqrt{2}}{2}$ times the distance from the corner); small epsilon in overlap and bound checks avoids numerical tangency.
  \item \emph{Variable radii:} optimal solutions for $n=26$ use circles of different sizes---larger circles at the center and corners, smaller ones filling gaps---with no uniform-radius assumption.
  \item \emph{$n=26$ target knowledge:} the benchmark target is sum-of-radii $\approx 2.635$ (AlphaEvolve); central hexagon plus outer layer arrangements work well; variable radii and corner optimization are the two most critical factors.
  \end{itemize}

\item \textbf{Optimization methodology (4 items).}
  \begin{itemize}[noitemsep,topsep=1pt]
  \item \emph{SLSQP constrained optimization:} maximize sum of radii subject to explicit no-overlap ($d_{ij} \geq r_i + r_j + \varepsilon$) and in-bounds constraints via \texttt{scipy.optimize.minimize}; warm-start from high-scoring nodes (score $\geq 2.2$).
  \item \emph{Multi-start strategy:} 3--5 different initial configurations from the best database nodes; optimize each independently and keep the best; tight tolerances with \texttt{maxiter} 500--1000.
  \item \emph{Differential evolution:} use \texttt{scipy.optimize.differential\_evolution} for global refinement when local optimization plateaus; optionally refine radii only (lower-dimensional subspace), then polish with SLSQP.
  \item \emph{AlphaEvolve reference:} AlphaEvolve achieved 2.635 via constructor-plus-constrained-optimization; key insight is good geometric initialization followed by strict-constraint numerical refinement.
  \end{itemize}

\item \textbf{Engineering and troubleshooting guidelines (4 items).}
  \begin{itemize}[noitemsep,topsep=1pt]
  \item \emph{Incremental refinement:} do not rewrite the full program; make targeted changes to optimizer settings, constraint formulation, multi-start count, or add a dedicated refinement stage.
  \item \emph{Code structure:} keep construction and optimization as separate stages; use a high-scoring node's code as base and apply diff-style edits.
  \item \emph{Numerical stability:} use a small epsilon (e.g.\ $10^{-8}$) in all constraints to avoid numerical tangency; verify that no constraint is violated before reporting the sum of radii.
  \item \emph{Plateau-breaking checklist:} if stuck at $\approx 2.3$--$2.4$, try (1) different initial pattern from a high-scoring node, (2) increase \texttt{maxiter} to 500--1000, (3) switch to explicit inequality constraints instead of penalty terms, (4) increase multi-start count to 3--5, (5) run differential evolution then SLSQP polishing.
  \end{itemize}
\end{itemize}

\paragraph{Note on the framework comparison.}
OpenEvolve is commonly reported with a fast/slow model pair, where a second model can affect proposal quality and selection dynamics. In our aligned comparison, we instead instantiated both ASI-Evolve and OpenEvolve-style settings with the same backbone model in each run, so that the comparison focuses on framework design rather than auxiliary-model choice. We acknowledge that this simplification may introduce bias relative to the best possible OpenEvolve configuration, since a different auxiliary model could potentially improve its performance. However, using the same model on both sides avoids an additional confound: results would otherwise depend heavily on how the second model is chosen, and a poorly chosen auxiliary model could distort the comparison even more. For GEPA, we retained all official settings except model-related parameters. We set the score of invalid evaluation outcomes to 0; otherwise the system would record invalid nodes with anomalously high scores and distort evolution. All other GEPA configuration and evaluation logic is unchanged from the official release.